\documentclass[9.9pt,logo,copyright]{nvidiatechreport}
\usepackage[most]{tcolorbox}
\newcommand{\finding}[2]{
    \vspace{0.5em}
    \begin{tcolorbox}[
        colback=white!90!gray,
        colframe=teal!60!black,
        arc=5pt,
        boxsep=5pt,
        left=10pt,
        right=10pt,
        top=2pt,
        bottom=2pt,
        boxrule=0.8pt,
        drop shadow=gray!0!white,
        enhanced jigsaw
    ]
    \vspace{-0.15cm}
        \minisection{Key Finding #1:} #2
    \vspace{-0.1cm}
    \end{tcolorbox}
    \vspace{-0.1cm}
}

\usepackage{url}            %
\usepackage{nicefrac}       %
\usepackage{microtype}      %
\usepackage{graphicx}
\usepackage{amsmath}
\usepackage{amssymb}
\usepackage{booktabs}
\usepackage{xspace}
\usepackage{amsmath,amsfonts,amssymb,amsthm}
\usepackage[dvipsnames]{xcolor}
\usepackage{diagbox}
\usepackage{pifont}
\usepackage{svg}
\usepackage{enumitem}
\usepackage{physics}
\usepackage{mathtools}
\usepackage{algorithm}
\usepackage{algorithmic}
\usepackage{gensymb}
\usepackage{wrapfig}
\usepackage{subcaption}
\usepackage{multirow}
\usepackage{footmisc}
\usepackage{empheq}
\usepackage{cases}
\usepackage{adjustbox}
\usepackage{array}
\usepackage{tabu}
\usepackage{flushend}
\usepackage{multirow}
\usepackage{bm}
\usepackage{color, colortbl}
\usepackage[numbers]{natbib}
\usepackage{makecell}
\usepackage{capt-of,etoolbox}
\usepackage[nameinlink]{cleveref}

\title{
Scaling Vision Pre-Training to 4K Resolution
}

\author{Baifeng Shi\textsuperscript{1,2*} \qquad Boyi Li\textsuperscript{1,2} \qquad Han Cai\textsuperscript{2} \qquad Yao Lu\textsuperscript{2} \qquad Sifei Liu\textsuperscript{2} \qquad Marco Pavone\textsuperscript{2}  \qquad \qquad \qquad \qquad \qquad   Jan Kautz\textsuperscript{2} \qquad Song Han\textsuperscript{2} \qquad Trevor Darrell\textsuperscript{1} \qquad Pavlo Molchanov\textsuperscript{2} \qquad Hongxu Yin\textsuperscript{2} \\~\\ \textsuperscript{1}UC Berkeley \quad \textsuperscript{2}NVIDIA}

\begin{abstract}

    High-resolution perception of visual details is crucial for daily tasks. Current vision pre-training, however, is still limited to low resolutions (\eg, 378$\times$378 pixels) due to the quadratic cost of processing larger images. We introduce \textbf{\model} that scales CLIP-style vision pre-training to 4K resolution with a \textit{near-constant} cost. Instead of contrastive learning on global image representation, \model is pre-trained by selectively processing local regions and contrasting them with local detailed captions, enabling high-resolution representation learning with greatly reduced computational overhead. The pre-trained \model is able to both encode the global image at low resolution and selectively process local high-resolution regions based on their saliency or relevance to a text prompt. When applying \model to multi-modal LLMs (MLLMs), the resulting model, named \textbf{\vilamodel}, significantly improves high-resolution visual perception compared to baselines without high-resolution vision pre-training such as AnyRes and \stwo while using up to 4.3$\times$ fewer tokens. \model also unlocks appealing scaling properties of \vilamodel, including scaling up resolution for free and scaling up test-time compute for better performance. Compared to state of the arts, \model and \vilamodel outperform previous vision encoders (\eg, SigLIP2 and Perception Encoder) and MLLMs (\eg, NVILA and Qwen2.5-VL) respectively across multiple benchmarks and achieve better efficiency than latest token pruning approaches. Finally, we find current benchmarks do not require 4K-resolution perception, which motivates us to propose \textbf{\benchmark}, a new benchmark of image QA at 4K resolution, on which \vilamodel outperforms all previous MLLMs, including a 16.1\% improvement over GPT-4o and a 7.5\% improvement and 1.67$\times$ speedup over Qwen2.5-VL.

\vspace{5pt}

\textbf{Project page:} \hspace{1pt} \textbf{\url{https://nvlabs.github.io/PS3}}

\end{abstract}

\begin{document}

\makeatletter
\DeclareRobustCommand\onedot{\futurelet\@let@token\@onedot}
\def\@onedot{\ifx\@let@token.\else.\null\fi\xspace}

\def\eg{\textit{e.g.}\xspace} \def\Eg{E.g\onedot}
\def\ie{\textit{i.e.}\xspace} \def\Ie{I.e\onedot}
\def\vs{vs.\xspace}
\def\cf{c.f\onedot} \def\Cf{C.f\onedot}
\def\etc{etc\onedot} \def\vs{vs\onedot}
\def\wrt{w.r.t\onedot} \def\dof{d.o.f\onedot}
\def\etal{et al\onedot}
\def\viz{viz\onedot}

\def\blfootnote{\xdef\@thefnmark{}\@footnotetext}

\makeatother

\newcommand{\cmark}{\ding{51}}%
\newcommand{\xmark}{\ding{55}}%

\newcommand{\customfootnotetext}[2]{{%
  \renewcommand{\thefootnote}{#1}%
  \footnotetext[0]{#2}}}%

\def\rvx{{\mathbf{x}}}
\def\rvy{{\mathbf{y}}}
\def\vx{{\bm{x}}}
\def\vy{{\bm{y}}}
\def\vmu{{\bm{\mu}}}
\def\mSigma{{\bm{\Sigma}}}
\def\mI{{\bm{I}}}

\newcommand{\model}{PS3\xspace}
\newcommand{\modeltwok}{PS3$_{\text{1512}}$\xspace}
\newcommand{\modelfourk}{PS3$_{\text{3780}}$\xspace}
\newcommand{\vilamodel}{VILA-HD\xspace}
\newcommand{\benchmark}{4KPro\xspace}
\newcommand{\stwo}{{S$^2$}\xspace}
\newcommand{\stwobf}{{\textbf{S$\mathbf{^2}$}}\xspace}
\newcommand{\stwowrapper}{{S$^2$-Wrapper}\xspace}

\definecolor{darkgreen}{HTML}{054907}
\definecolor{darkgreen_teaser}{HTML}{105611}
\definecolor{lightgreen}{HTML}{9CCBB8}
\definecolor{lightred}{HTML}{E3242B}
\definecolor{lightorange}{HTML}{ED7D31}
\definecolor{nvidiagreen}{HTML}{76B900}

\newcommand{\yin}[1]{\textcolor{cyan}{\textbf{[Yin: #1]}}}
\newcommand{\PM}[1]{\textcolor{orange}{\textbf{[Pavlo: #1]}}}

\newcommand\minisection[1]{\vspace{1.3mm}\noindent \textbf{#1}}

\newcommand{\bs}[1]{\textcolor{red}{\textbf{[Baifeng: #1]}}}
\newcommand{\drawio}[1]{{\color{Gray}{\bf drawio: #1}}}
\newcommand{\han}[1]{{\color{blue}{\bf Han: #1}}}

\newcommand{\PreserveBackslash}[1]{\let\temp=\\#1\let\\=\temp}
\newcolumntype{C}[1]{>{\PreserveBackslash\centering}p{#1}}
\newcolumntype{R}[1]{>{\PreserveBackslash\raggedleft}p{#1}}
\newcolumntype{L}[1]{>{\PreserveBackslash\raggedright}p{#1}}

\newcommand{\ver}[1]{\rotatebox[origin=l]{90}{#1}}
\newcommand{\multirowver}[1]{\rotatebox[origin=l]{90}{\parbox{1.5cm}{#1}}}

\maketitle

\abscontent

\section{Introduction}

Vision models with large-scale pre-training~\cite{oquab2023dinov2,radford2021learning,he2022masked,dosovitskiy2020image} have been the workhorses for both fundamental vision tasks~\cite{kirillov2023segment,yang2024depth} and numerous downstream applications~\cite{radosavovic2023real,ramesh2022hierarchical,juneja2024dino}. Notably, CLIP-style vision pre-training (\ie, vision-language contrastive learning) such as CLIP~\cite{radford2021learning} and SigLIP~\cite{zhai2023sigmoid} have driven significant advancements in multi-modal large language models (MLLMs) by providing general-purpose language-aligned visual understanding in real-world tasks~\cite{liu2024nvila,liu2023visual,li2024llava}.

\begin{figure}
    \centering
    \includegraphics[width=1\linewidth]{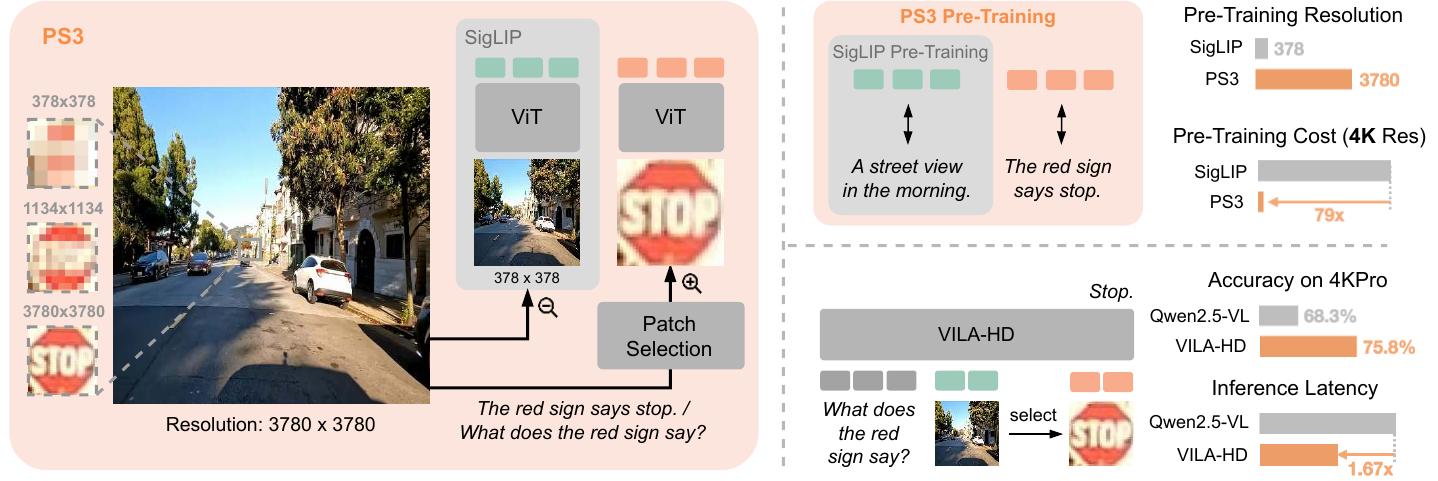}
    \caption{\textbf{Left:} Regular vision models such as SigLIP processes images at a low resolution (\eg, 378 $\times$ 378 pixels), which is not enough for many daily tasks such as spotting the stop sign while driving. In contrast, \model is able to both encode low-res features and efficiently process high-res information of 4K-resolution images via top-down patch selection, \ie, selectively processing relevant patches based on any text prompt. \textbf{Top Right:} SigLIP is pre-trained by contrasting global vision features and global captions, which is costly for high-resolution images. \model is pre-trained with additional contrast between local high-res features with local captions, enabling pre-training at 4K resolution with 79$\times$ less cost than SigLIP. \textbf{Bottom Right:} VILA-HD with \model as the vision encoder is able to select high-res regions to process based on the user prompt. VILA-HD outperforms state-of-the-art MLLMs such as Qwen2.5-VL~\cite{bai2025qwen2} by 7.5\% on the proposed 4KPro benchmark while achieving 1.67$\times$ speedup. }
    \label{fig:teaser}
\end{figure}

However, modern vision models including CLIP and SigLIP have one defect: they are \textit{pre-trained with low resolution only}. Visual perception at high resolution (\eg, 4K resolution) is essential in many real-world scenarios such as spotting the stop sign while driving (Figure~\ref{fig:teaser}(Left)). On the other hand, SigLIP, for example, is only pre-trained with a maximum resolution of 378$\times$378~\cite{zhai2023sigmoid}, making it incapable of perceiving visual details and thus unsuitable for assisting humans in everyday tasks. Existing methods propose to run pre-trained vision models at higher resolution in a training-free manner for downstream tasks~\cite{shi2025we,liu2024llavanext,chen2024far}. However, this prevents the model from leveraging large-scale pre-training data to learn high-quality high-resolution perception, resulting in suboptimal performance~\cite{shi2025we}.

What blocks the current vision pre-training from scaling to higher resolution? The computational cost. The compute spent by the vision model grows quadratically for CNNs and quartically for ViTs with increasing image resolution, making it even infeasible to pre-train over 1K resolution~\cite{oquab2023dinov2,zhai2023sigmoid}. 

In this work, we introduce \textbf{P}re-training with \textbf{S}cale-\textbf{S}elective \textbf{S}caling, or \textbf{\model}, that scales CLIP-style pre-training to \textit{4K resolution} with a \textit{near-constant cost}. The key insight is that, instead of contrasting between global images and captions for the whole high-res image, it suffices to contrast between local regions and local captions to learn detailed feature extraction in high-resolution images. For example, in Figure~\ref{fig:teaser}(Left, Top Right), to learn to recognize the text on the stop sign, the model only needs to extract the high-resolution feature around the local region of the text and align it with the detailed description about the region. This is analogous to top-down selection mechanism in human vision~\cite{li2014understanding,carrasco2011visual}, \ie, one usually focuses on a small portion of the scene that is relevant to the high-level task (\eg, spotting the stop sign). In this way, the model enjoys greatly reduced computational cost by being scale-selective, \ie, selectively processing a small region at fine-grained scale. By disentangling the region size from the image resolution, we are able to scale \model pre-training to 4K resolution with a near-constant cost, reducing the pre-training compute by 79$\times$ compared to global contrastive learning of SigLIP (Figure~\ref{fig:teaser}(Top Right)).

The success of \model pre-training hinges on addressing three challenges: \textit{data}, \textit{model}, and \textit{algorithm}. First, since the low-resolution image-text pairs used for CLIP pre-training is not suitable for \model pre-training, we collect 75M images with up to 4K resolution and build an automatic pipeline to curate 282M pairs of detailed captions and bounding boxes of salient local regions in the images. Second, we design a vision model that can not only extract low-resolution global features, but also select local patches based on image saliency or text queries and process high-resolution details of the patches. Third, we design an algorithm that pre-trains high-resolution perception through contrastive loss between local regions and local captions and pre-trains patch selection with supervision from the curated bounding boxes.

We show \model enables high-quality and efficient high-resolution perception in multi-modal LLMs (MLLMs). Specifically, we train a modern MLLM~\cite{liu2024nvila} using pre-trained \model as the vision encoder. The resulting MLLM, named \textbf{\vilamodel}, is capable of capturing the global image at low resolution and extracting high-resolution details in the local regions selected based on the user prompt. Evaluated on seven benchmarks that require high-res perception, \vilamodel significantly improves the performance over baselines that use either the original low-res SigLIP or approaches such as \stwo~\cite{shi2025we} and AnyRes~\cite{chen2024far,liu2024llavanext} that scale up the resolution of SigLIP without high-resolution vision pre-training, while using 4.3$\times$ fewer tokens compared to the AnyRes baseline. Additionally, \model unlocks several intriguing scaling properties of \vilamodel, for example, scaling up the resolution without extra cost by selecting a constant number of high-res patches and trading more compute for higher performance at test time by selecting larger high-res regions. Comparing to the state of the arts, \model outperforms previous vision encoders such as SigLIP2~\cite{tschannen2025siglip} and Perception Encoder~\cite{bolya2025perception} on 23 benchmarks under a controlled setting, and with a more advanced MLLM training recipe, \vilamodel is able to surpass state-of-the-art MLLMs such as NVILA~\cite{liu2024nvila} and Qwen2.5-VL~\cite{bai2025qwen2} on various benchmarks and achieve superior efficiency and performance over latest token pruning approaches~\cite{bolya2022token,chen2025image,yang2024visionzip}. 

Despite the superior performance of \model, we find most existing benchmarks do not actually require 4K resolution. Therefore, we introduce \textbf{\benchmark}, a benchmark that evaluates visual perception at 4K resolution in four professional use cases including autonomous vehicle, household, gaming, and UI understanding. For each category, 4KPro contains image QA pairs where each question can only be answered under 4K resolution. On 4KPro, \model shows a significant improvement of 15\% over \stwo and AnyRes baselines and achieves state-of-the-art results compared to both proprietary and open-sourced MLLMs including GPT-4o~\cite{hurst2024gpt} and Qwen2.5-VL~\cite{bai2025qwen2} in terms of both performance and efficiency (Figure~\ref{fig:teaser}(Bottom Right)).

\section{\model: Vision Pre-Training at 4K Resolution}

Based on the paradigm of contrastive language-image pre-training (CLIP)~\cite{radford2021learning} which optimizes a contrastive loss between global images and global captions, we propose \model which instead optimizes the contrast loss between local regions and detailed captions about the regions (Figure~\ref{fig:teaser}(Upper Right)). In this way, the model efficiently learns language-aligned detailed representation by being scale-selective, \ie, only processing the selected regions at a fine-grained scale. This detaches the computational cost from the global image size, allowing us to scale up to ultra-high image resolution during pre-training by controlling the size of the local regions.

The scale-selective pre-training requires a redesign of data, model, and algorithm. We first collect 75M high-resolution images with 282M pairs of bounding boxes and captions of salient local regions (Section~\ref{sec:pretrain_data_curation}). We then design the architecture of \model that can both encode low-resolution global images and select local high-resolution patches to process based on image saliency or their relevance to a text prompt (Section~\ref{sec:pretrain_model_design}). We finally pre-train \model jointly with localized contrastive loss for high-res perception and box supervision for patch selection (Section~\ref{sec:pretrain_algorithm}).

\subsection{Pre-Training Data of \model}
\label{sec:pretrain_data_curation}

\begin{figure}
    \centering
    \begin{minipage}[b]{0.47\textwidth}
        \includegraphics[width=\linewidth]{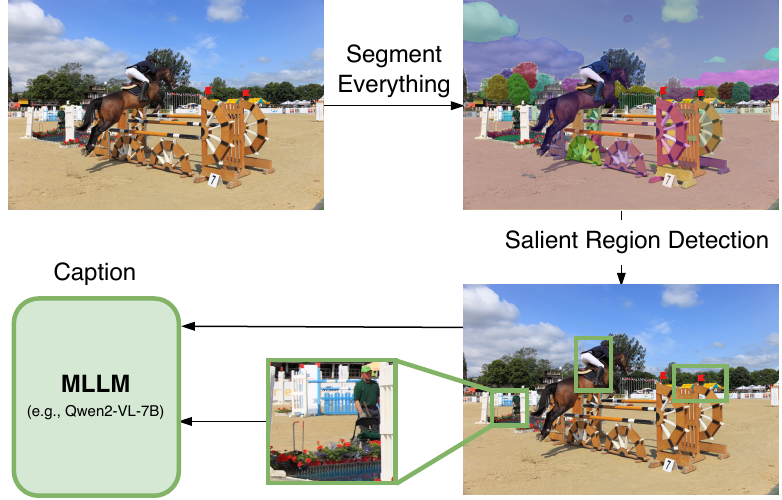}
        \caption{\textbf{Curation of bounding boxes and captions of salient regions in the pre-training data.} For each high-resolution image, we segment all the masks, detect salient regions with small or dense masks, and use an MLLM to generate captions about the local regions.%
        }
        \label{fig:data_curation}
    \end{minipage}
    \hfill
    \begin{minipage}[b]{0.47\textwidth}
        \includegraphics[width=\linewidth]{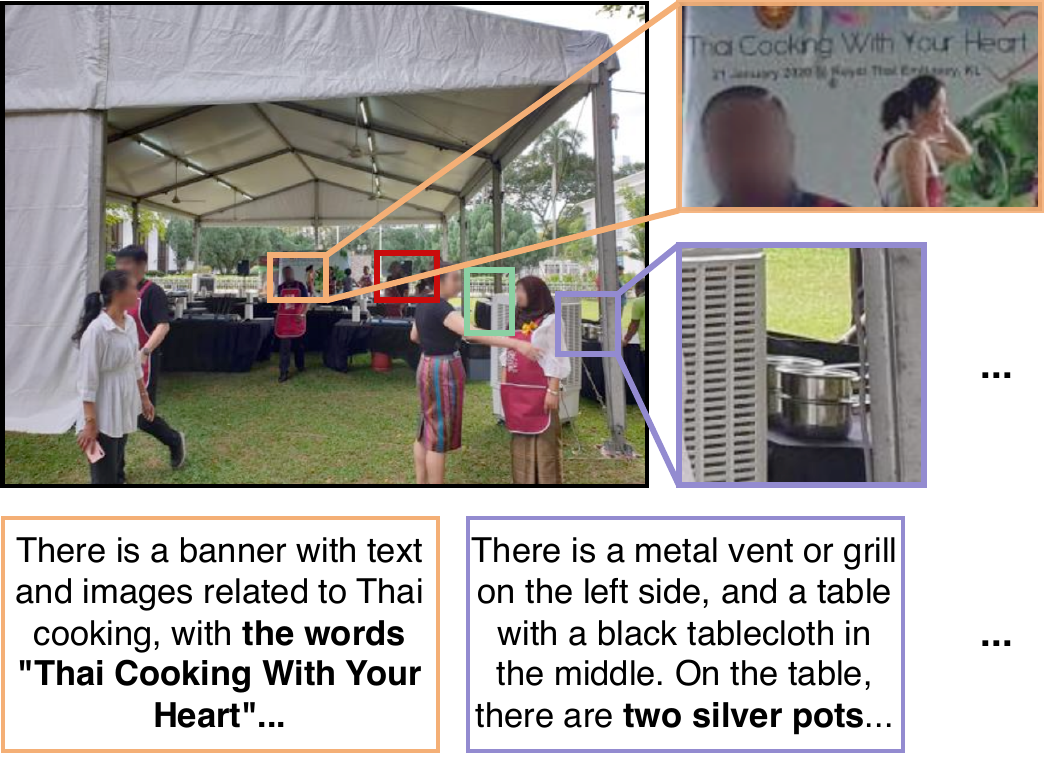}
        \caption{\textbf{Pre-training data example.} Each instance contains an image with resolution up to 4K, bounding boxes of the salient regions in the image, and captions about details in the regions such as text or small objects. }
        \label{fig:data_example}
    \end{minipage}
    \vspace{-0.5em}
\end{figure}

To learn fine-grained perception in high-res images through contrastive loss between local regions and captions, we need to collect high-res images together with bounding boxes and captions of local regions in each image. We need to make sure the local regions contain rich details in order for the model to learn fine-grained representation. In this work, we collect 75M high-res images with 282M pairs of bounding boxes and captions for both natural images and document images. An example of the pre-training data is shown in Figure~\ref{fig:data_example} and more examples of natural and document images can be found in Figure \ref{appendix_fig:pretrain_data_example_sam}-\ref{appendix_fig:pretrain_data_example_idl} in the Appendix. The curation of each component of the data is detailed below.

\minisection{High-resolution images.} We collect 75M images with 1K - 4K resolution. These include 38M natural images from DataComp~\cite{gadre2024datacomp} and SA-1B~\cite{kirillov2023segment} and 37M document images from IDL~\cite{biten2022ocr} and PDFA~\cite{pdfa}.

\minisection{Local captions and bounding boxes of salient regions for natural images.} 
We propose a pipeline of first detecting salient regions and then generating local captions (Figure~\ref{fig:data_curation}). For saliency detection, inspired by recent work on segmenting everything in an image~\cite{kirillov2023segment,zhang2024efficientvit,wang2024segment}, we first use EfficientViT-SAM~\cite{zhang2024efficientvit} to generate segmentation masks of the whole image, and then select local regions containing small or dense masks as salient regions. The intuition is that a local region should contain small or cluttered objects in order to have rich details. The saliency detection algorithm is explained in more details in the Appendix~\ref{appendix_sec:data_curation}. For local captions, we use an off-the-shelf MLLM (\eg, Qwen2-VL~\cite{wang2024qwen2}) as an captioner. Specifically, we zoom in and crop the local region, send it along with the global image to the MLLM, and let it describe the local region based on the global context. This results in 3 - 4 pairs of local captions and bounding boxes per image and 134M pairs in total, with an average box size around 400$\times$400. 

\minisection{Local captions and bounding boxes for documents.} Both IDL~\cite{biten2022ocr} and PDFA~\cite{pdfa} provide bounding boxes and OCR results of sentences or words in each PDF. Therefore, we simply sample from these bounding boxes and use the corresponding OCR results as the local captions. We generate 148M pairs of boxes and captions with an average box size around 50$\times$400.

\minisection{Global captions.} Our pre-training also uses global captions (see Section~\ref{sec:pretrain_algorithm}). We use the same MLLM captioner to generate global captions for natural images. For document images we do not use any global captions.

\subsection{Model Design of \model}
\label{sec:pretrain_model_design}

\begin{figure*}
    \centering
    \includegraphics[width=1\linewidth]{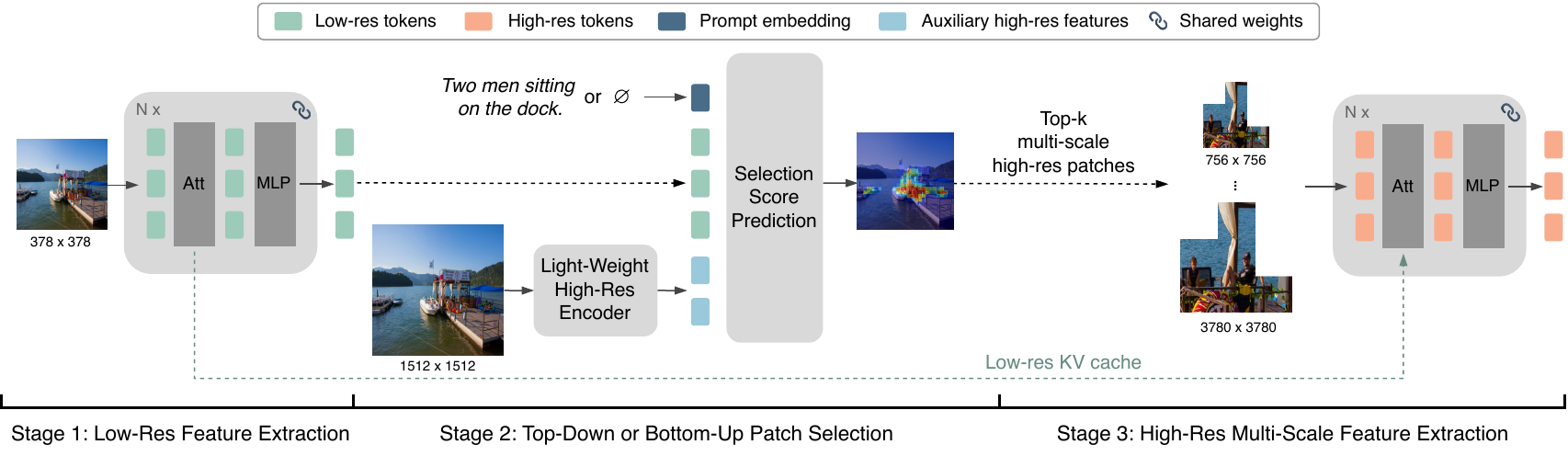}
    \caption{\textbf{Model architecture of \model.} The model consists of 3 stages. In Stage 1, the model encodes global low-resolution features. In Stage 2, based on the low-resolution features as well as auxiliary high-resolution features extracted by a light-weight encoder, the model selects local regions that are either relevant to a text prompt (top-down selection) or salient by themselves (bottom-up selection). In Stage 3, the model processes multi-scale high-res patches from the selected regions with the same encoder from Stage 1. KV cache from the low-res tokens in Stage 1 is added to the self-attention layers to provide a global context for local high-res encoding. 
    }
    \label{fig:model_arch}
\end{figure*}

We design the model such that given a high-res image, it can 1) extract low-res global features, 2) select local regions based on saliency or their relevance to an input text prompt, and 3) extract high-res features of the selected regions. The whole model can be divided into three stages corresponding to these three capabilities respectively (Figure~\ref{fig:model_arch}). The design of each stage is detailed below.

\minisection{Stage 1: Low-res feature extraction.} We use the same vision transformer (ViT) architecture as SigLIP-SO400M~\cite{zhai2023sigmoid} to extract low-res features. The image is resized to 378$\times$378 which corresponds to 27$\times$27 output tokens. %

\minisection{Stage 2: Top-down or bottom-up patch selection.} In this stage, the model selects important regions either based on their relevance to a text prompt (\ie, top-down selection) or based on the saliency of the region itself (\ie, bottom-up selection)~\cite{li2014understanding,carrasco2011visual}. See Figure~\ref{fig:patch_selection_example}(Left) for examples of such selection. To achieve this, the model predicts a selection score for each spatial position of the image by calculating the cosine similarity between the low-res visual features (from Stage 1) and the embedding of the prompt. The prompt embedding is either the text embedding in the case of top-down selection or a constant learnable vector in the case of bottom-up selection, following~\cite{shi2023top}. The text embedding comes from the text encoder in our contrastive pre-training.

The selection score is calculated with low-res features only, making it infeasible to locate fine-grained details. To alleviate this issue, we predict additional high-res selection score following the same process but with auxiliary high-res features extracted by a light-weight encoder. The light-weight encoder is a ConvNeXt~\cite{liu2022convnet} model with only 3 blocks and extracts features at 1512 resolution. The high-res and low-res selection scores are then interpolated to the same size and averaged as the final score.

\minisection{Stage 3: High-res multi-scale feature extraction.} Stage 3 consists of a few key steps. \textbf{1)~Selecting top-$k$ multi-scale high-res patches.} The model first resizes the high-res image to a set of pre-defined scales and patchifies each. For example, we use three scales of 756$\times$756, 1512$\times$1512, and 3780$\times$3780 for a maximum resolution of 4K. Each is then patchified to 54$\times$54, 108$\times$108, and 270$\times$270 patches, respectively. The selection score from Stage 2 is also interpolated into the same each size.
Then for each scale, top-$k$ patches with the highest score are selected. Note that $k$ can vary for different scales. During pre-training, we set $k$ for each scale to be proportional to the total number of patches at that scale, \eg, $k$ equals to 80, 320, 2000 for scales of 756$\times$756, 1512$\times$1512, and 3780$\times$3780. This ensures the exact same image region is selected for each scale. However, the user can flexibly decide which scale to select more patches by setting different $k$ for different scales, which we find benefits different downstream tasks (Section~\ref{sec:ablation_trafe_off_image_scales}). 
\textbf{2)~Scale-aware positional embedding.} After embedding each selected patch with the same patch embedding module as in Stage 1, we then add positional embedding for each token by interpolating the original low-res positional embedding and selecting the positional embeddings that correspond to the selected patches. On top of that, we add a scale-specific positional embedding to tokens from each scale such that the ViT is able to discriminate tokens from the same spatial position but different scales. \textbf{3)~High-res feature extraction with low-res KV cache.} The selected tokens from different scales are gathered and simultaneously processed by the same ViT as in Stage 1. To make the local high-res features aware of the global visual context, we augment the keys and values in the self-attention layers with the keys and values from the corresponding layer in the low-res feature extraction stage, similar to the KV cache in modern LLMs~\cite{radford2019language}. The effect of each step above is studied in Section~\ref{sec:ablation_model_algorithm}.

\subsection{Pre-Training Algorithm of \model}
\label{sec:pretrain_algorithm}

\begin{figure}
    \centering
    \includegraphics[width=\linewidth]{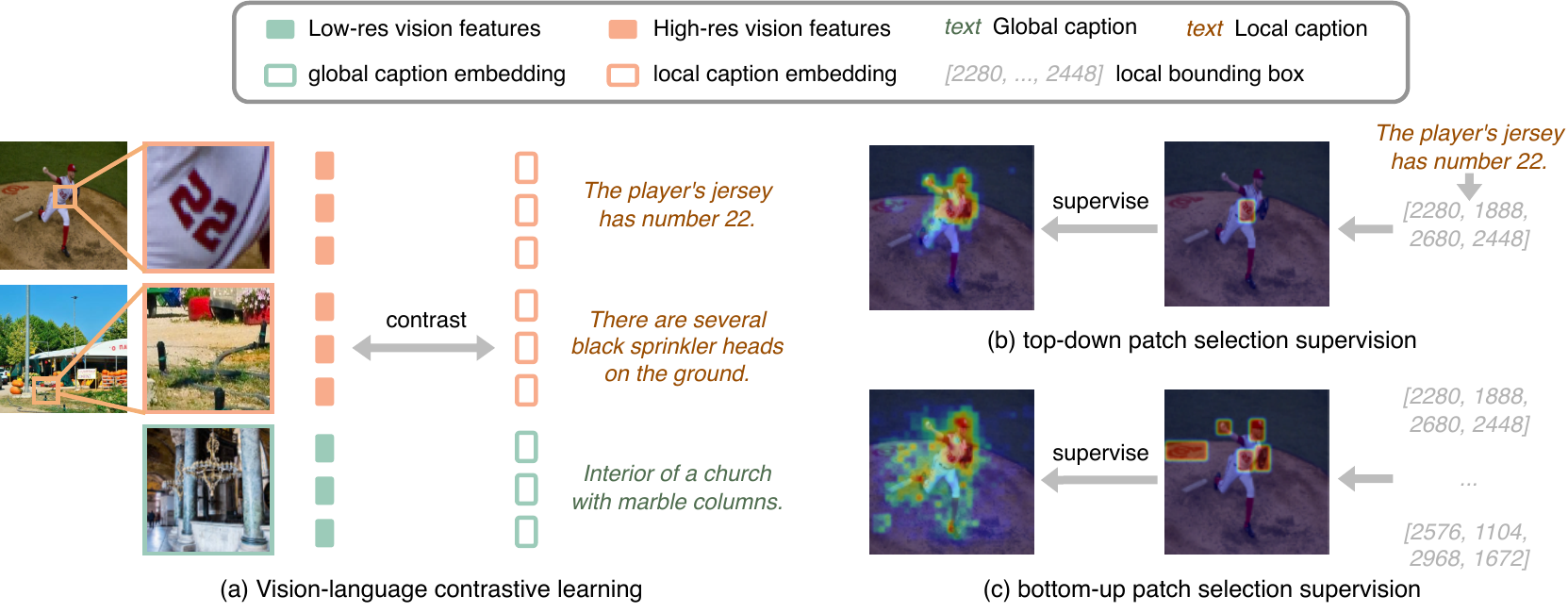}
    \caption{\textbf{Pre-training algorithm of \model.} \textbf{(a)} During training, \model extracts the high-res features from the labeled local regions and contrasts them with embeddings of the local captions. To maintain the low-res feature quality, we also mix pairs of low-res features and global caption embedding in each batch. Both high-res and low-res features are extracted in the same way as Figure~\ref{fig:model_arch}. \textbf{(b)} The top-down patch selection score is supervised by ground-truth score map generated from the bounding box corresponding to the local caption. \textbf{(c)} The supervision for bottom-up selection is similar to top-down selection, except that the ground-truth selection score is generated from all the labeled bounding boxes of the image.}
    \label{fig:pretrain_algorithm}
\end{figure}

\begin{figure}
    \centering
    \includegraphics[width=0.95\linewidth]{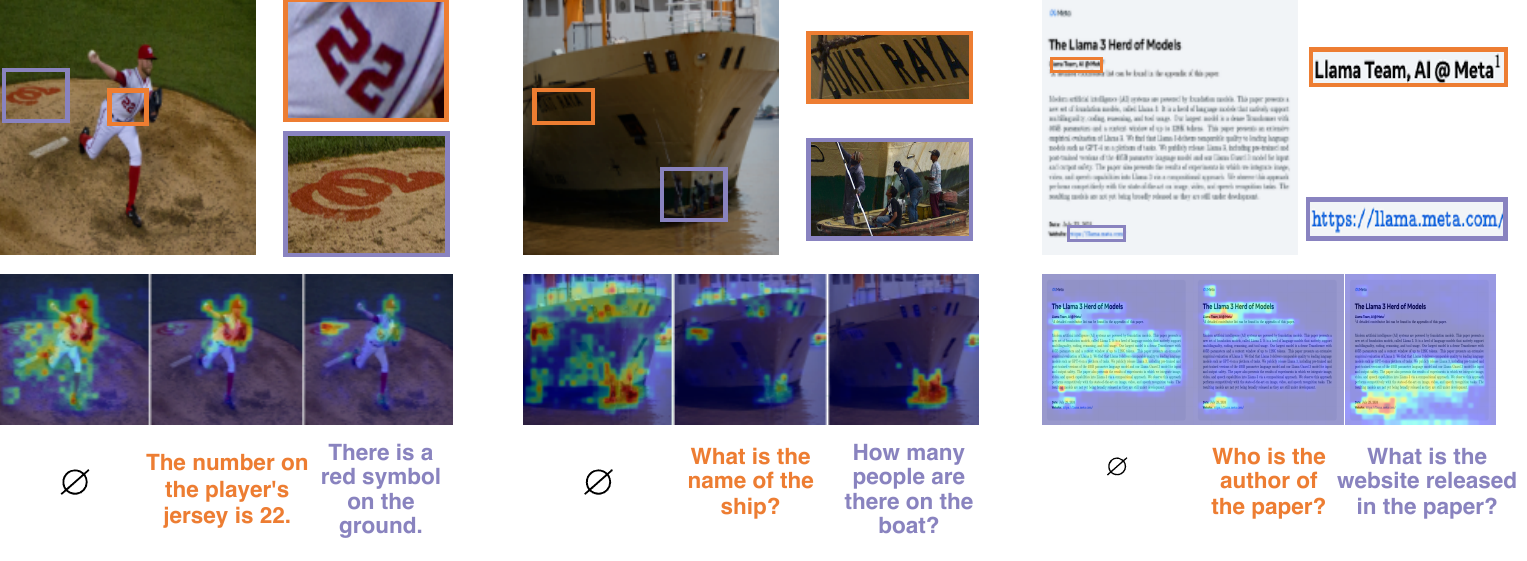}
    \vspace{-1em}
    \caption{\textbf{Qualitative examples of patch selection.} \textit{Left}: \model is pre-trained to perform bottom-up selection based on image saliency (denoted by $\varnothing$) or top-down selection based on local captions. The selection process is detailed in Figure~\ref{fig:model_arch} and Section~\ref{sec:pretrain_model_design}. \textit{Middle} \& \textit{Right}: We fine-tune \model with MLLM to select patches based on questions about local regions (Figure~\ref{fig:ps3_for_vila} and Section~\ref{sec:ps3_for_vila}).  %
    }
    \label{fig:patch_selection_example}
\end{figure}

\model is pre-trained to jointly learn 1) detailed visual representation through localized contrastive loss and 2) top-down and bottom-up patch selection from box supervision.

\minisection{Learning high-res visual representation.} As illustrated in Figure~\ref{fig:pretrain_algorithm}(a), given the paired data of high-res images and detailed local captions, we use \model to extract the high-res features of the local regions that are relevant to the local captions as described in Section~\ref{sec:pretrain_model_design}, extract text embedding of the local captions using a text encoder, and optimize a contrastive loss between the high-res visual features and the text embeddings. The total number of selected high-res patches for each image is limited to 2560 during pre-training for efficiency, while one can choose to select more tokens for downstream applications (Section~\ref{sec:ps3_for_vila}). We use the same sigmoid contrastive loss as in SigLIP~\cite{zhai2023sigmoid}. Both the ViT backbone in \model and the text encoder are initialized with the pre-trained SigLIP.

There are several key designs in the contrastive pre-training. The effect of each design is studied in Section~\ref{sec:ablation}. \textbf{1) Using ground-truth selection score for patch selection.} Normally \model selects patches based on the local caption. However, in pre-training, to avoid inaccuracy in the selection score predicted by the model which may lead to selecting irrelevant regions, we use selection score generated from the ground-truth bounding box. This is similar to Teacher Forcing~\cite{williams1989learning} in training recurrent neural networks, where the ground truth of an intermediate prediction is given in order to better train the consequent prediction of the model. The ground truth selection score is generated by setting the score inside the box to 1 and others to 0.  \textbf{2) Pooling only tokens in the ground-truth boxes.} SigLIP uses attention pooling to compress all the output tokens into one for contrastive loss. When a box contains fewer patches than the pre-set $k$, the model will select patches outside the box as well. Pooling both tokens inside and outside the box results in aligning irrelevant visual features to the text embedding in contrastive loss. To avoid this, we constrain the attention pooling to only tokens inside the box.  \textbf{3) Mixing global and local contrast.} We empirically find that optimizing contrastive loss only between local regions and captions can degrade the quality of global low-res representations. To this end, we mix global and local contrast, \ie, each batch contains both pairs of global low-resolution features and global caption embeddings and pairs of local high-resolution features and local caption embeddings.  \textbf{4) Avoiding intra-image contrast.} Since we have multiple local boxes and captions for each high-res image, there is a chance that one batch contains multiple local regions from the same image. It can be problematic to contrast between different regions of the same image if those regions are visually similar~\cite{chen2024contrastive}. We make sure each image only appears once in a batch to avoid intra-image contrast. The effect of each design is studied in Section~\ref{sec:ablation_model_algorithm}.

\minisection{Learning top-down and bottom-up patch selection.} As illustrated in Figure~\ref{fig:pretrain_algorithm}(b), we use box supervision to learn top-down and bottom-up patch selection. During pre-training, the model predicts both the bottom-up selection score (\ie, without any text prompt) and the top-down selection score based on the local caption for each image. To learn top-down patch selection, we supervise the selection score with the ground-truth bounding box that corresponds to the local caption. Specifically, we treat it as a binary semantic segmentation problem, where the predicted likelihood of binary classification is the selection score at each position and the ground truth label is 1 inside the ground-truth box and 0 outside. A position-wise cross entropy loss as well as a DICE loss~\cite{sudre2017generalised} is then optimized between the selection score map and the ground-truth map. For bottom-up selection, since we already label the bounding boxes of salient regions in each image, we directly use these boxes to generate ground truth segmentation map by setting the label of a position to 1 as long as it belongs to one of the boxes. We then supervise the bottom-up selection score using the same loss functions as above. Figure~\ref{fig:patch_selection_example}(Left) shows examples of the learned top-down and bottom-up patch selection.

\section{\vilamodel: Enabling High-Resolution MLLM with \model}

We apply \model to MLLMs to enhance their high-resolution perception capability. Specifically, we propose \vilamodel, an MLLM with \model as the vision encoder that shows suprior performance and efficiency in processing images of up to 4K resolution. In the following we introduce the model design (Section~\ref{sec:ps3_for_vila}) and training data (Section~\ref{sec:train_vila_w_ps3}) of \vilamodel.

\subsection{Building \vilamodel with \model}
\label{sec:ps3_for_vila}

\begin{figure}
    \centering
    \vspace{-0.5em}
    \includegraphics[width=1\linewidth]{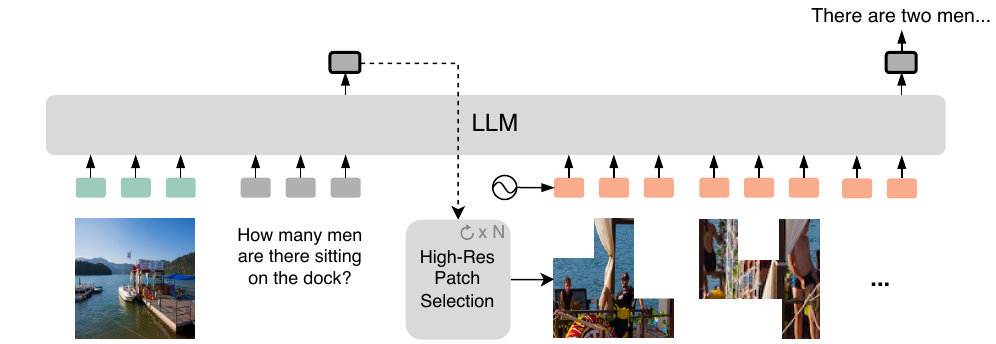}
    \caption{\textbf{Model design of \vilamodel.} For any input image and text prompt, \vilamodel first extracts the low-res image features using \model and sends them along with the text tokens to the LLM. The last-layer embedding of the last token is used to select high-res patches in \model, whose features are then extracted by \model, added with additional positional embedding, and sent to the LLM. Although \model only processes at most 2560 high-res patches at a time, one can run the patch selection and high-res feature extraction for N times (N can be an arbitrary number) to encode up to 2560$\times$N high-res patches.}
    \label{fig:ps3_for_vila}
\end{figure}

\vilamodel's model architecture is based on NVILA~\cite{liu2024nvila}. NVILA is a LLaVA-style~\cite{liu2023visual} MLLM with SigLIP~\cite{zhai2023sigmoid} as the vision encoder, and we replace the vision encoder with \model to build \vilamodel. The model design of \vilamodel is illustrated in Figure~\ref{fig:ps3_for_vila}. We first extract the global low-res features following Stage 1 of \model and send them along with the text tokens to LLM. We then select high-res patches in either a bottom-up or a top-down way. Bottom-up selection is exactly the same as in pre-training. For top-down selection, since we need to select regions that can help answer the user's question, instead of using the text embedding from the SigLIP text encoder as the prompt, we use the latent embedding of the last token in the user's text input from the last layer of LLM as the prompt embedding. This is inspired by LISA~\cite{lai2024lisa} which uses the same embedding for reasoning segmentation. Finally, the selected high-res patches are encoded by Stage 3 of \model and sent to LLM after the text tokens, from which the following text generation resumes. We also add an additional positional embedding to the high-res features before sending them to LLM such that LLM is aware of the spatial positions of the selected patches, similar to the scale-aware positional embedding Stage 3 of \model. Note that while the number of selected high-res patches is limited to 2560 during pre-training, one can select an arbitrary number of patches when applying to MLLMs by running patch selection and high-res feature extraction for multiple times. For example, to select 3840 high-res patches, one can first select the top 2560 patches to process in Stage 3, and then select another top 1280 patches among the rest of the unselected patches and process them in Stage 3. The multiple runs of Stage 3 are independent from each other. Then all the high-res tokens are gathered together and sent to LLM.

\subsection{Fine-Tuning Data of \vilamodel}
\label{sec:train_vila_w_ps3}

\begin{figure}
    \centering
    \includegraphics[width=\linewidth]{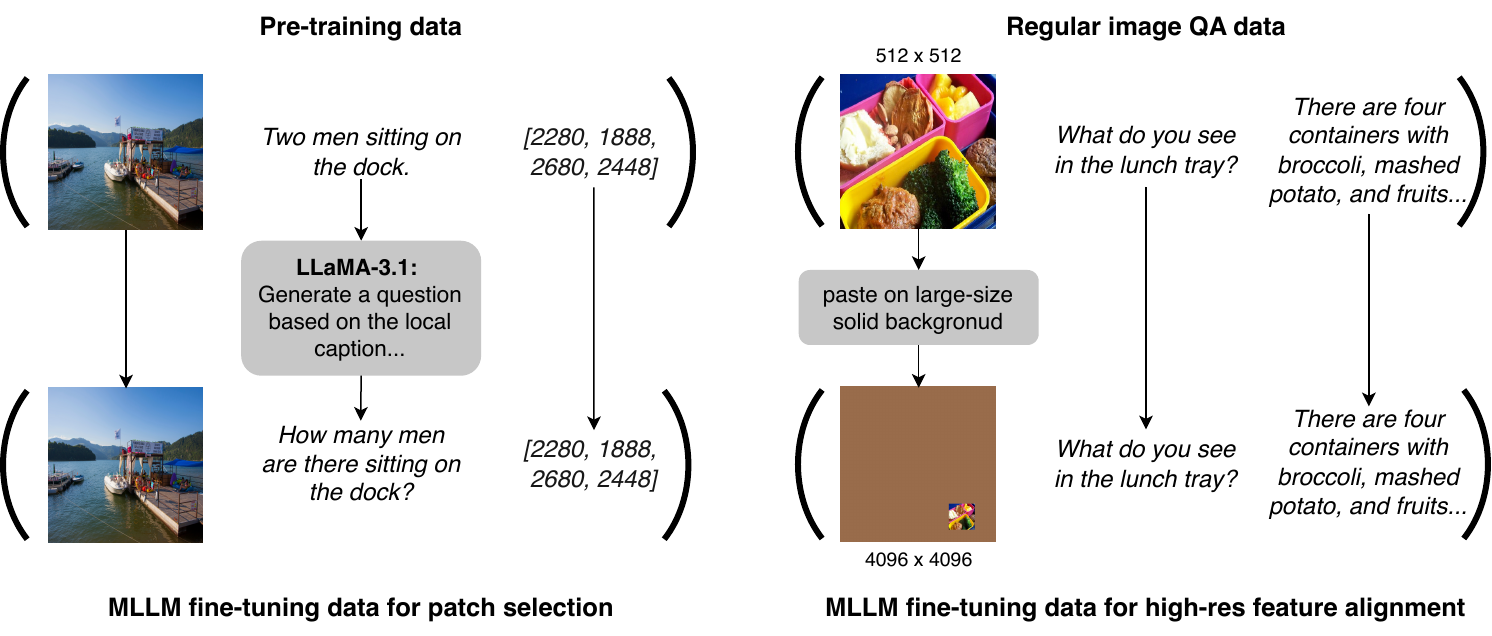}
    \caption{\textbf{Additional fine-tuning data for MLLMs with \model.} \textit{Left:} To fine-tune top-down patch selection, we generate data with pairs of high-res image, question about a local region, and the bounding box of the region. This is generated by taking the \model pre-training data and retargeting the local captions into questions using LLaMA-3.1. \textit{Right:} To align the \model high-res features to the LLM text space, it requires fine-tuning data that contains QA pairs on high-res images. We generate this by taking regular low-res image QA data and pasting the image onto a large-size background to get the new high-res image while the question and answer are inherited.}
    \label{fig:sft_data}
\end{figure}

Except for regular fine-tuning data for training MLLMs in previous work~\cite{liu2024nvila,li2024llava}, we collect additional data for 1) learning top-down patch selection in \vilamodel and 2) aligning the high-res features from \model to the text space of \vilamodel. Each type of data is introduced below and Figure~\ref{fig:sft_data} illustrates the curation of the data.

\minisection{Fine-tuning data for patch selection.} When training \vilamodel, we jointly fine-tune the top-down patch selection since it uses different prompt embeddings compared to pre-training. For this purpose, we collect data of high-res images paired with \textit{questions} about local regions as well as their bounding boxes. As shown in Figure~\ref{fig:sft_data}(a) This is achieved by directly sampling images and bounding boxes from the pre-training data (Section~\ref{sec:pretrain_data_curation}) and then automatically generate questions based on the local captions using LLaMA-3.1~\cite{dubey2024llama}. We generate 500k data samples from the natural images and document images in the pre-training dataset respectively. To check the quality of the generated data, we randomly sample 500 examples and manually check whether the generated questions correctly corresponds to the original captions and local regions, and find 95.8\% of the questions are accurate. During training, we run \vilamodel with \model on this data to perform top-down patch selection based on the questions, and then supervise the patch selection following the same objective as in Section~\ref{sec:pretrain_algorithm}. Note that this data does not have answers to the questions so we do not optimize the next-token-prediction loss on this data. See Figure~\ref{fig:patch_selection_example}(Middle \& Right) for visualization of the learned patch selection.

\minisection{Fine-tuning data for high-res vision feature alignment.} Since most of the existing fine-tuning data for MLLMs only contains low-resolution images, the high-res vision features from \model, especially the features extracted at 4K resolution, are not properly aligned to the LLM's text space if they are only trained on low-resolution data. To solve this problem, we collect 225k VQA data at up to 4K resolution. The data curation pipeline is shown in Figure~\ref{fig:sft_data}(b). Specifically, we take existing low-res VQA data, paste the images onto a solid-color background image with larger size such as 4096$\times$4096, and use the new synthesized image with the original question as the high-res VQA data. By training on this data, \vilamodel is forced to utilize the high-res features from \model to answer the question because the informative region in the synthesized image is small, \eg, a 256$\times$256 region in a 4K image. During training, we mix this data with other fine-tuning data. We find this simple approach works reasonably well in our empirical studies (Section~\ref{sec:ablation_data}).

\section{Scaling Properties of PS3}
\label{sec:scaling_property}

\begin{figure}
    \centering
    \includegraphics[width=0.97\linewidth]{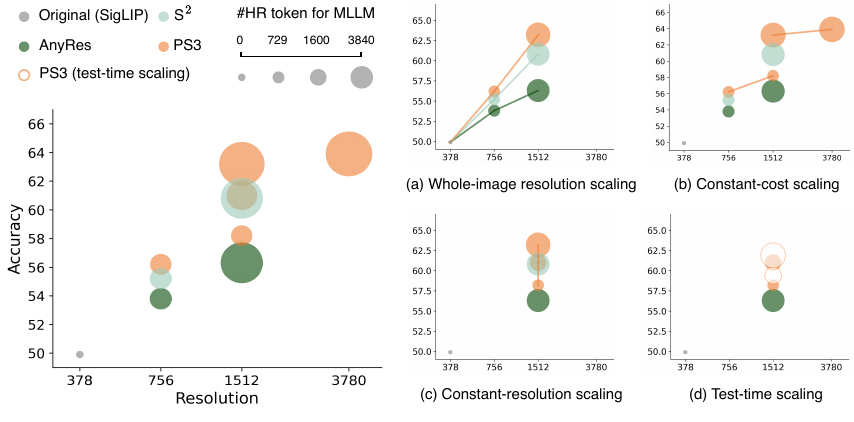}
    \vspace{-1em}
    \caption{\textbf{Scaling properties of \model on \vilamodel.} \textbf{(Left)} Overall results. We report average performance of the MLLM on seven benchmarks under different maximum input resolution. The size of each data point indicates the number of high-res vision tokens input to the LLM. \textbf{(a)} When selecting all high-res patches for MLLM, the performance of \model scales better with the resolution than the baselines without high-resolution pre-training. \textbf{(b)} \model is able to process higher resolution and improve performance while selecting a fixed number of high-res patches for MLLM. \textbf{(c)} Within the same resolution, \model is able to trade compute for performance by selecting more high-res patches. \textbf{(d)} \model can select more high-res patches at test time even if its selects a fixed number of high-res patches during MLLM training.
    }
    \label{fig:scaling_curve}
\end{figure}

In this section, we evaluate how well the performance of \model scales with the pre-training resolution. Specifically, we pre-train \model at different resolutions and compare the performance of \vilamodel with the \model encoders as well as baseline vision encoders using a fixed MLLM training recipe. We pre-train \model at three resolutions of 756, 1512, and 3780, where for each resolution we extract multi-scale high-res features at scales of (756), (756, 1512), and (756, 1512, 3780), respectively. We show four types of scaling: \textbf{1) Whole-image resolution scaling.} \model selects all the high-res patches at each resolution for \vilamodel. This is to compare the high-res feature quality of \model with the baselines that also process all high-res patches (see \textit{Experiment settings} below). \textbf{2) Constant-cost scaling.} \model selects a constant number of high-res patches when scaling the resolution for \vilamodel. This evaluates if performance scales ``for free'', \ie, by maintaining a constant MLLM training and inference cost (note that the pre-training cost is already near-constant). \textbf{3) Constant-resolution scaling.} At the same resolution, \model selects increasingly more high-res patches for \vilamodel. This evaluates if we gain benefits from selecting more patches in downstream without changing the pre-training and fine-tuning resolution. \textbf{4) Test-time scaling.} Similar to 3), but we increase the number of high-res patches at \textit{test} time. The scaling curves are shown in Figure~\ref{fig:scaling_curve} and more detailed quantitative results are reported in Table~\ref{tab:scaling}. %

\minisection{Experiment settings.} \model is initialized with SigLIP-SO400M~\cite{zhai2023sigmoid} before pre-training. Please see detailed pre-training setting in Appendix~\ref{appendix_sec:pretrain_detail}. \vilamodel is trained using a subset of the training data in NVILA. Please see the detailed training recipe of \vilamodel in Appendix~\ref{appendix_sec:mllm_training_setting}. For evaluation, we report average accuracy on seven resolution-sensitive benchmarks: TextVQA~\cite{singh2019towards}, ChartQA~\cite{masry2022chartqa}, DocVQA~\cite{mathew2021docvqa}, InfoVQA~\cite{mathew2022infographicvqa}, OCRBench~\cite{liu2023hidden}, V$^\ast$Bench~\cite{wu2024v}, and RealWorldQA~\cite{realworldqa}. We compare \model to the original SigLIP as well as two baselines, AnyRes~\cite{chen2024far,liu2024llavanext} and \stwo~\cite{shi2025we}, that run SigLIP at larger resolution in a training-free way by splitting large images into smaller tiles. For AnyRes baselines, we set the maximum number of tiles to 4 for 756 resolution and 9 for 1512 resolution. For \stwo baselines, instead of resizing the feature maps from different scales to the smallest scale as in the original algorithm, we resize them to the largest scale to get the best performance. For \model, the number of selected patches for multiple image scales at test time might vary between benchmarks to obtain the best performance (Section~\ref{sec:ablation_trafe_off_image_scales}). Please see Appendix~\ref{appendix_sec:mllm_training_setting} for the detailed evaluation setting.

\begin{table}
\caption{\textbf{Full results of scaling properties of \model on \vilamodel.} \textit{\#HR Token} is the number of high-res tokens input to the MLLM. \textit{Select (Train)} is the percentage of high-res patches \model selects when training the MLLMs and \textit{Select (Test)} is the selection ratio at test time. Note the number of HR tokens is usually 1/4 of the number of selected patches due to the 2$\times$2 downsampling connector in NVILA~\cite{liu2024nvila}. $^\dagger$\textit{\#HR tokens} for AnyRes depends on the input resolution and we report the maximum number of tokens.}
    \label{tab:scaling}
    \vspace{-0.5em}
    \centering
    \begin{small}
    \begin{tabular}{llp{0.055\textwidth}p{0.05\textwidth}p{0.05\textwidth}cccccccc}
        \toprule
        \makecell[l]{Vision \\ Encoder} & \makecell[l]{Max \\ Res.} & \makecell[l]{\#HR \\ Token} & \makecell[l]{Select \\ (Train)} & \makecell[l]{Select \\ (Test)} & \makecell{Text \\ VQA} & \makecell{Chart \\ QA} & \makecell{Doc \\ VQA} & \makecell{Info \\ VQA} & \makecell{OCR \\ Bench} & \makecell{V$^\ast$ \\ Bench}  & \makecell{Real \\ World}  & \textit{Avg} \\ 
        \midrule
        SigLIP~\cite{zhai2023sigmoid} & 378 & 0 & - & - & 62.3 & 56.6 & 51.9 & 30.7 & 387 & 51.8 & 57.1 & \textit{49.9} \\
        \midrule
        AnyRes~\cite{liu2024llavanext} & 756 & 784$^\dagger$ & - & - & 65.3 & 58.0 & 60.6 & 32.7 & 416 & 59.2 & 59.1 & \textit{53.8} \\
        \stwo~\cite{shi2025we} & 756 & 729 & - & - & 65.9 & 65.5 & 63.0 & 32.3 & 471 & 53.1 & 59.6 & \textit{55.2} \\
        \rowcolor{lightorange!20} & 756 & 320 & 44\% & 44\% & 66.7 & 62.8 & 62.6 & 33.1 & 460 & 56.3 & 61.7 & \textit{55.6} \\
        \rowcolor{lightorange!20} \multirow{-2}{*}{\model} & 756 & 729 & 100\% & 100\% & 66.8 & 63.5 & 64.6 & 33.9 & 462 & 56.5 & 61.7 & \textit{56.2} \\
        \midrule
        AnyRes~\cite{liu2024llavanext} & 1512 & 3136$^\dagger$ & - & - & 67.4 & 58.4 & 67.9 & 34.1 & 468 & 60.2 & 59.0 & \textit{56.3} \\
        \stwo~\cite{shi2025we} & 1512 & 2916 & - & - & 66.1 & 71.0 & 78.3 & 41.1 & 526 & 55.2 & 61.0 & \textit{60.8} \\
        \rowcolor{lightorange!20} & 1512 & 729 & 20\% & 20\% & 67.3 & 64.7 & 66.5 & 34.8 & 505 & 60.7 & 62.6 & \textit{58.2}  \\
        \rowcolor{lightorange!20} & 1512 & 1600 & 20\% & 44\% & 67.7 & 65.9 & 70.7 & 35.7 & 515 & 62.0 & 62.6 & \textit{59.4} \\
        \rowcolor{lightorange!20} & 1512 & 1600 & 44\% & 44\% & 68.4 & 68.0 & 74.5 & 37.3 & 509 & 63.1 & 65.0 & \textit{61.0}  \\
        \rowcolor{lightorange!20} & 1512 & 3645 & 44\% & 100\% & 68.4 & 68.0 & 76.5 & 39.4 & 522 & 66.7 & 62.0 & \textit{61.9} \\
        \rowcolor{lightorange!20} \multirow{-5}{*}{\model} & 1512 & 3645 & 100\% & 100\% & 69.3 & 71.1 & 79.4 & 41.3 & 534 & 64.0 & 63.8 & \textit{63.2} \\
        \midrule
        AnyRes~\cite{liu2024llavanext} & 3780 & 19600$^\dagger$ & - & - & \multicolumn{8}{c}{OOM} \\
        \stwo~\cite{shi2025we} & 3780 & 18225 & - & - & \multicolumn{8}{c}{OOM} \\
        \rowcolor{lightorange!20} \model & 3780 & 3840 & 18\% & 18\% & 69.8 & 70.9 & 79.1 & 40.5 & 543 & 67.8 & 64.7 & \textit{63.9} \\
        \bottomrule
    \end{tabular}
    \end{small}
\end{table}

\subsection{Whole-Image Resolution Scaling}
\label{sec:whole_image_resolution_scaling}

We first scales up the resolution of \model and make it select all the high-res patches at each resolution for \vilamodel. Results are shown in Figure~\ref{fig:scaling_curve}(a). We can see that the effect of scaling up resolution is significant, where \model at 1512 resolution improves 14.2\% over SigLIP baseline. Compared to AnyRes and \stwo, \model shows consistent improvements across different resolution while using a similar number of high-res tokens. For example, at 1512 resolution, \model improves by 2.4\% over \stwo and 6.9\% over AnyRes which is commonly used by modern MLLMs~\cite{li2024llava,chen2024far}. Since all the methods are processing the whole high-res image, the advantage of \model mainly comes from the improved high-res feature quality which is brought by our high-res pre-training. Note that we cannot scale up to 4K resolution while selecting all the high-res patches due to the drastic computational cost, although it is achievable with constant-cost scaling in Section~\ref{sec:constant_cost_scaling}.

\subsection{Constant-Cost Scaling}
\label{sec:constant_cost_scaling}

Scaling up resolution comes at a cost of quadratically increasing number of tokens in Section~\ref{sec:whole_image_resolution_scaling}. However, for \model, higher resolution is still beneficial even when selecting a constant number of patches (Figure~\ref{fig:scaling_curve}(b)). For example, using 729 (20\%) high-res tokens with 1512 resolution improves the accuracy by 2\% over using 729 (100\%) tokens with 756 resolution (Table~\ref{tab:scaling}). This is because at 756 resolution, not all patches are relevant to the user's questions, and by scaling up the resolution to 1512, even it selects the same number of patches, there are fewer irrelevant 756-resolution patches but more relevant 1512-resolution patches being selected, thanks to our top-down selection mechanism. Comparing to AnyRes, \model achieves 58.2\% accuracy at resolution of 1512 through constant-cost scaling, improving over AnyRes by 2\% accuracy while using 5$\times$ fewer tokens. Notably, while neither AnyRes or \stwo can scale to 4K resolution, \model is able to scale up to 4K resolution with a constant training and inference cost, where \model achieves 63.9\% accuracy, improving over \stwo at 1512-resolution by 3.1\% while using a similar number of tokens. Specifically, the accuracy on V$^\ast$Bench improves from 64.0\% to 67.8\% by scaling to 4K resolution, which shows that 4K resolution is important for such detailed visual perception. We observe an even larger advantage for tasks that require 4K resolution (Section~\ref{sec:4kpro_main_results}).

\subsection{Constant-Resolution Scaling}
\label{sec:constant_resolution_scaling}

Pre-trained at a fixed resolution, \model can flexibly select different number of patches for \vilamodel to trade compute for performance. As shown by Figure~\ref{fig:scaling_curve}(c), selecting more patches at 1512 resolution consistently improves performance. By increasing the number of high-res tokens from 729 (20\%) to 1600 (44\%), \model is able to outperform \stwo with only half the number of high-res tokens (Table~\ref{tab:scaling}). Further increasing to 3645 (100\%) tokens boosts the performance by another 2.2\%. We observe that the improvement is especially significant for tasks that has dense visual information and requires processing most parts of the image such as document understanding in DocVQA. On the other hand, tasks such as TextVQA and RealWorldQA which only require understanding of local regions only have minor performance drops (2.0\% and 1.2\%, respectively) when we select only 20\% of the high-res patches. In practice, \model allows one to select different number of high-res patches based on the computational budget and the requirements for high-res information density in different tasks.

\subsection{Test-Time Scaling}
\label{sec:test_time_scaling}

Constant-resolution scaling is still valid at test time, \ie, we select a fixed number of high-res patches during training but select more at test time. As shown in Figure~\ref{fig:scaling_curve}(d), at 1512 resolution, we can train with 20\% high-res patches and test with 44\% patches, which improves the accuracy by 1.2\%. Similarly, training with 40\% patches but testing with 100\% patches gives an improvement of 0.9\%. Note that scaling at test time still performs worse than training time, which is possibly because 1) the MLLM receives lower-quality supervision if seeing fewer patches at training time, and 2) the LLM is not adapted to longer context length during training.

\vspace{0.5em}
\finding{1}{Scaling up the pre-training resolution of \model significantly improves performance of downstream MLLM over baselines without high-res pre-training. This is achievable with near-constant pre-training cost and even with no extra downstream training and inference cost by selecting a fixed number of high-res patches. \model can also flexibly select different number of patches to trade training or test-time compute for better performance.}

\begin{figure}[h!]
\centering
  \includegraphics[width=0.6\linewidth]{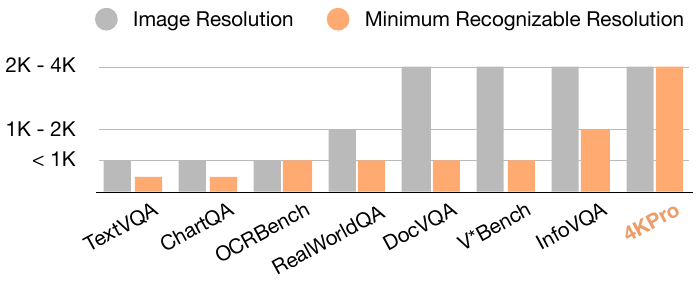}
  \vspace{-0.5em}
  \caption{\textbf{Image resolution and MRR of different benchmarks.} Existing benchmarks contain high-res images but the resolution required to answer the questions (MRR) is mostly under 1K. In contrast, 4KPro contains questions only solvable at 4K resolution.}
  \label{fig:mrr}
\end{figure}

\section{4KPro: Benchmarking \model at 4K Resolution}
\label{sec:4kpro}

Despite the suprior performance of \model and \vilamodel on existing benchmarks as shown in the previous section, we find those benchmarks do not actually require high resolution visual perception, especially 4K-resolution perception, even though some of them contain high-resolution images. To verify this, we examine the minimum recognizable resolution (MRR) of the existing benchmarks, \ie, the minimum resolution required to answer the questions. We calculate the MRR by randomly sampling examples from each benchmark, manually checking the minimum resolution (4K, 2K, or 1K) under which the visual details are clear enough to 
answer the question for each example, and averaging the minimum resolutions of different samples for each benchmark. As shown in Figure~\ref{fig:mrr}, we can see for all previous benchmarks their MRR is lower than the average image resolutions, which means these benchmarks don't require visual perception at the same resolution as the images. Specifically, even though benchmarks like DocVQA and V$^\ast$Bench contain images at 4K resolution, the MRR is mostly around 1K. InfoVQA has the highest MRR of 2K, although it is solely focused on infographic understanding. 

To effectively evaluate 4K-resolution perception in real-world tasks, we introduce 4KPro, an image QA benchmark with MRR of 4K from four professional use cases including Autonomous Vehicle, Household, Gaming, and UI Understanding. Each QA pair is in the form of multi-choice problem with four options. Examples of 4KPro are shown in Figure~\ref{fig:qualitative_4kpro}, where we can see that all the questions cannot be accurately answered without 4K resolution. We detail the data curation process in Appendix~\ref{appendix_sec:4kpro_curation}.

\minisection{Experiment settings.} The training setting is the same as Section~\ref{sec:scaling_property} where we pre-train \model at resolutions of 756, 1512, and 3780, train \vilamodel with each PS3 encoder and evaluate on 4KPro. When comparing with state-of-the-art MLLMs, we use SigLIP2-SO400M~\cite{tschannen2025siglip} as initialization during \model pre-training and evaluate \vilamodel with \model pre-trained at 1512 and 3780 resolutions, which we denote by \vilamodel-8B-\model-1.5K-SigLIP2 and \vilamodel-8B-\model-4K-SigLIP2, and we use the full training recipe in NVILA-8B~\cite{liu2024nvila} with differences detailed in Appendix~\ref{appendix_sec:mllm_training_setting}.

\subsection{Main Results}
\label{sec:4kpro_main_results}

\begin{figure}[h!]
    \centering
    \vspace{2em}
    \includegraphics[width=1\linewidth]{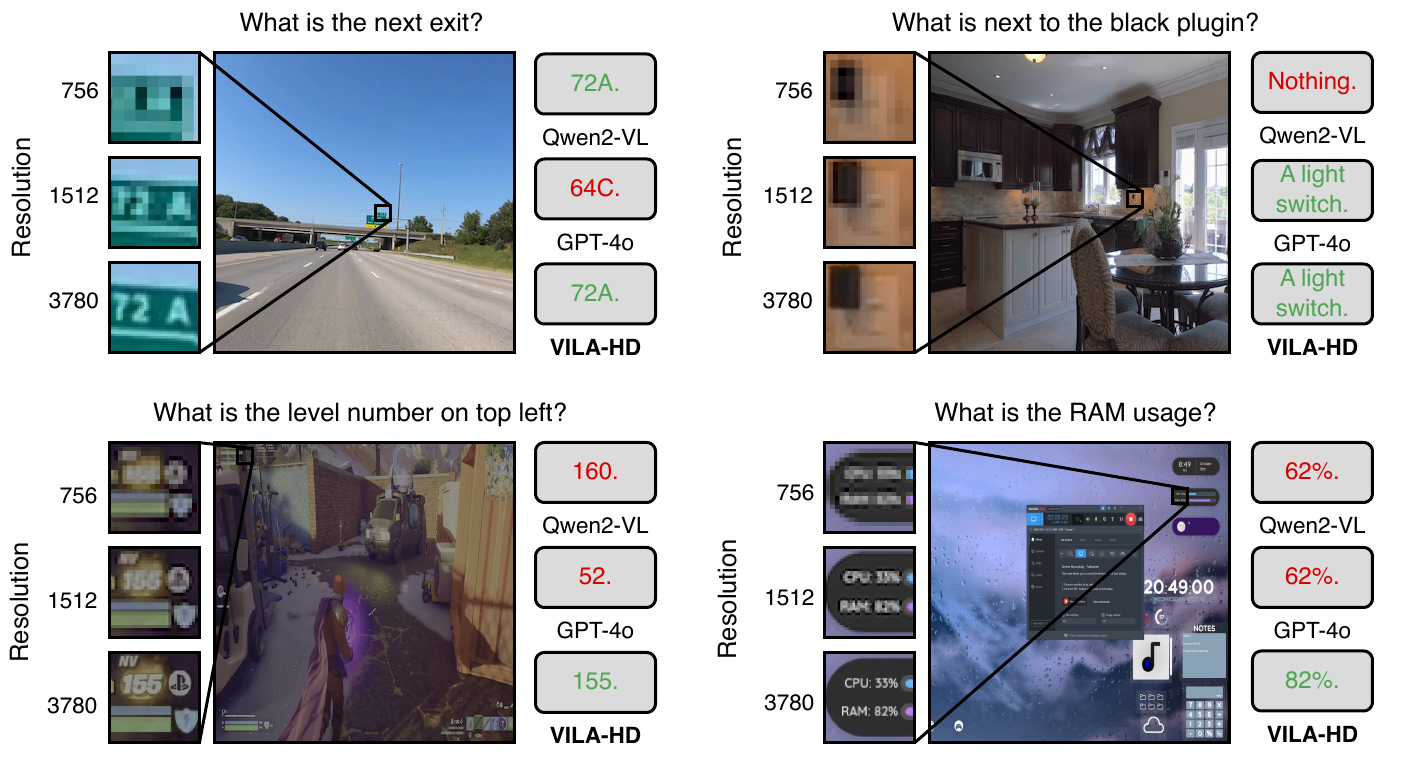}
    \vspace{-1em}
    \caption{\textbf{Examples from 4KPro and comparison of different models}. Each example corresponds to one out of four categories (Autonomous Vehicle, Household, Gaming, and UI Understanding) and each question can only be answered without ambiguity under 4K resolution. VILA-\model improves the accuracy over the state-of-the-art MLLMs such as GPT-4o and Qwen2-VL.}
    \label{fig:qualitative_4kpro}
    \vspace{2em}
\end{figure}

\begin{figure}[h!]
    \centering
    \begin{minipage}[b]{0.45\textwidth}
    
        \caption{\textbf{Scaling properties of \model on 4KPro.} \model shows consistently improved performance by scaling to 4K resolution and achieves further improvements by scaling up number of high-res tokens at test time, greatly outperforming the baselines.}
        \label{fig:scaling_curve_4kpro}
        \includegraphics[width=0.983\linewidth]{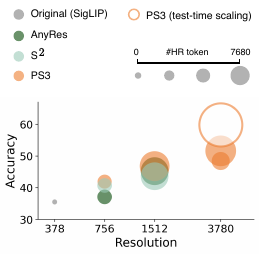}
        
    \end{minipage}
    \hfill
    \begin{minipage}[b]{0.5\textwidth}
    
        \captionof{table}{\textbf{Comparing \vilamodel-8B-\model-1.5K-SigLIP2 and \vilamodel-8B-\model-4K-SigLIP2 to state-of-the-art MLLMs on 4KPro.} %
        $^\dagger$We limit the maximum number of pixels to 4,000,000 to maintain a reasonable latency.}
        \label{tab:sota_4kpro}
        \centering
        \begin{small}
        \begin{tabular}{lccc}
        \toprule
           Model & Select & Latency & Acc.   \\
        \midrule
           GPT-4o~\cite{hurst2024gpt} & - & - & 59.7 \\
           Claude 3.5 Sonnet~\cite{claude35sonnet} & - & - & 29.0 \\
           Gemini-1.5-Pro~\cite{team2024gemini} & - & - & 59.7 \\
        \midrule
           NVILA-8B~\cite{liu2024nvila} & - & 0.82s & 58.1 \\
           Cambrian-1-8B~\cite{tong2024cambrian} & - & 2.78s & 50.0 \\
           InternVL2-8B~\cite{internvl2} & - & 1.65s & 58.1 \\
           IXC-2.5-7B~\cite{zhang2024internlm} & - & 2.11s & 32.3 \\
           LLaVA-OneVision-7B~\cite{li2024llava} & - & 1.75s & 67.7 \\
           InternVL3-8B~\cite{zhu2025internvl3} & - & 2.78s & 61.3 \\
           Qwen2-VL-7B-Instruct$^\dagger$~\cite{wang2024qwen2} & - & 3.61s & 71.0 \\
           Qwen2.5-VL-7B-Instruct$^\dagger$~\cite{bai2025qwen2} & - & 2.98s & 68.3 \\
           
        \midrule
           \rowcolor{lightorange!20} \vilamodel-1.5K & 33\% & 0.46s & 59.7 \\ 
           \rowcolor{lightorange!20} \vilamodel-1.5K & 100\% & 1.20s & 67.7 \\ 
        \midrule
           \rowcolor{lightorange!20} \vilamodel-4K & 18\% & 1.22s & 71.0 \\
           \rowcolor{lightorange!20} \vilamodel-4K & 35\% & 1.78s & \textbf{75.8} \\
           
        \bottomrule
        \end{tabular}
        \end{small}
        
    \end{minipage}
    \vspace{0em}
\end{figure}

\minisection{Scaling properties of \model.} We evaluate the performance of \vilamodel when scaling up the resolution of \model. Results are shown in Figure~\ref{fig:scaling_curve_4kpro}. We can see \model outperforms other baselines at resolution of 756 and 1512, although by a relatively small margin, which implies 4KPro is genarlly unsolvable with resolution lower than 4K. However, while it is infeasible to scale to 4K resolution for the baselines, \model enables 4K resolution perception by selecting the same number of high-res patches as 1512 resolution, using the same constant-cost scaling scheme as in Section~\ref{sec:scaling_property}. This improves the performance by 4.8\%. Taking a step further, we can double the number of high-res patches at test time to boost the performance by another 8.2\%. This achieves an improvement of $\sim$15\% over AnyRes at 1512 resolution, which is more significant than the improvements we get by scaling to 4K resolution in Section~\ref{sec:scaling_property}, indicating that 4K-resolution pre-training is most helpful for downstream tasks requiring ultra high-res perception. On the other direction, we can also shrink the number of patches by 3$\times$, achieving 48.4\% accuracy which is 3.2\% higher than AnyRes at 1512 resolution while using 2.5$\times$ fewer tokens. 

\minisection{Comparison to state of the arts.} We compare the performance of \vilamodel-8B-\model-1.5K-SigLIP2 and \vilamodel-8B-\model-4K-SigLIP2 with other proprietary or open-source MLLMs (Table~\ref{tab:sota_4kpro}), where we abbreviated the two models as \vilamodel-1.5K and \vilamodel-4K. All the proprietary MLLMs do not perform well, with the best accuracy at 59.7\% for GPT-4o and Gemini-1.5-Pro. We find part of the errors for proprietary MLLMs are caused by erroneous instruction following or refusing to answer the question when unable to recognize the details. For open-source models, most achieve accuray around 50\% - 60\%. Qwen2-VL-7B-Instruct performs the best, reaching an accuracy of 71.0\%, although at a cost of larger latency than other models due to processing the full high-resolution images in its vision encoder. \vilamodel-4K, on the other hand, achieves 75.8\% accuracy at 3780 resolution when selecting 35\% patches, which improves over both the proprietary models and the state-of-the-art open-source model (Qwen2-VL) while having a lower latency thanks to the top-down patch selection mechanism. \vilamodel-4K also significantly improves over \vilamodel-1.5K by 8.1\%. Note that the performance under 1.5K resolution is still better than random guess (25\%) even though the visual details are not clear, which is probably because the blurry details still provide some clues for an informed guess. By selecting fewer patches (\eg, 18\%), \vilamodel-4K still maintains superior performance of 71.0\% while enjoying only 1/3 of the latency comparing to Qwen2-VL. We also show qualitative results in Figure~\ref{fig:qualitative_4kpro}. We can see that compared to state-of-the-art MLLMs, \vilamodel with \model is able to accurately recognize extremely fine-grained details under 4K resolution, \eg, tiny text on highway signs or monitor screens and small light switch on a distant wall.

\finding{2}{Most of previous benchmarks do not require perception at 4K resolution despite the images at 4K resolution, and on 4KPro benchmark that does require 4K resolution perception, \vilamodel with \model significantly improves over previous MLLMs in both accuracy and efficiency.}

\section{Comparing \model and \vilamodel to State of the Arts}
\label{sec:sota}

In this section, we compare \model and \vilamodel with other state-of-the-art models. Specifically, we compare the visual representation quality of \model with state-of-the-art vision encoders~\cite{tschannen2025siglip,Ranzinger_2024_CVPR,bolya2025perception,fini2024multimodal,chen2024expanding}, compare the performance of \vilamodel with state-of-the-art MLLMs~\cite{hurst2024gpt,claude35sonnet,team2024gemini,lin2023vila,tong2024cambrian,liu2024nvila,zhang2024mm1,li2024llava,dong2024internlm,zhang2024internlm,internvl2,wang2024qwen2}, compare the efficiency of patch selection in \model and \vilamodel with other MLLM token pruning methods~\cite{bolya2022token,chen2025image,yang2024visionzip}, and also verify the generalizability and effectiveness of \model pre-training pipeline on different state-of-the-art vision encoders~\cite{zhai2023sigmoid,Ranzinger_2024_CVPR}.

\subsection{Comparing \model to State-Of-The-Art Vision Encoders}
\label{sec:sota_vision_encoder}

\renewcommand{\arraystretch}{1.2}
\begin{table}
\caption{\textbf{Comparison to state-of-the-art vision encoders.} We train an MLLM on top of each vision encoder with the same recipe and compare the performance on various types of benchmarks including high-resolution VQA, VQA over text, documents, and charts, and general VQA. We compare \textbf{\model-1.5K-SigLIP2} and \textbf{PS3$_{\text{Lang}}$-1.5K-SigLIP2} (abbreviated as \model and PS3$_{\text{Lang}}$ here) with previous state-of-the-art vision encoders~\cite{tschannen2025siglip,Ranzinger_2024_CVPR,bolya2025perception,fini2024multimodal,chen2024expanding}. For a fair comparison, we use similar ViT sizes (300M - 400M) for every encoder and use \stwo to scale all the vision encoders (except for \model which natively supports high resolution) to a similar high resolution round 1.5K. We compare the vision encoders with and without LLM alignment separately.}
    \label{tab:sota_vision_encoder}
    \vspace{-0.5em}
    \centering
    \begin{small}
    \begin{tabular}{p{0.23\textwidth}cccccccc}
        \toprule
        & \multicolumn{5}{c}{\cellcolor{Gray!20}\textit{\textcolor{Gray}{\textbf{w/o LLM alignment}}}} & \multicolumn{3}{c}{\cellcolor{Gray}\textit{\textcolor{Gray!10}{\textbf{w/ LLM alignment}}}} \\
        
        Benchmark & SigLIP2 & C-RADIOv2 & PE$_{\text{Core}}$ & AIMv2 & \cellcolor{lightorange!20} \model & InternViT2.5 & PE$_{\text{Lang}}$ & \cellcolor{lightorange!20} PS3$_{\text{Lang}}$ \\
        
        \midrule
        
        \textcolor{Gray!50}{\textit{\textbf{High-Resolution VQA}}} &  &  &  &  & \cellcolor{lightorange!20} &  &  & \cellcolor{lightorange!20} \\

        4KPro & 37.1 & 32.3 & 46.8 & 43.5 & \cellcolor{lightorange!20} \textbf{50.0} & 43.5 & 45.2 & \cellcolor{lightorange!20} \textbf{53.2} \\

        HRBench$_{\text{4K}}$~\cite{wang2025divide} & 55.0 & 51.8 & 56.1 & 51.5 & \cellcolor{lightorange!20} \textbf{58.5} & 54.6 & 52.0 & \cellcolor{lightorange!20} \textbf{62.9} \\
        
        HRBench$_{\text{8K}}$~\cite{wang2025divide} & 49.9 & 44.8 & 50.3 & 47.6 & \cellcolor{lightorange!20} \textbf{52.5} & 47.3 & 49.9 & \cellcolor{lightorange!20} \textbf{50.5} \\
        
        MME-RealWorld~\cite{zhang2024mme} & 43.7 & 40.7 & \textbf{44.1} & 43.4 & \cellcolor{lightorange!20} 42.1 & \textbf{44.9} & 43.5 & \cellcolor{lightorange!20} 40.4 \\
        
        V$^\ast$~\cite{wu2024v} & 58.8 & 59.7 & \textbf{64.9} & 63.1 & \cellcolor{lightorange!20} 64.2 & 57.4 & 61.1 & \cellcolor{lightorange!20} \textbf{62.7} \\
        
        \textit{Avg} & 48.9 & 45.9 & 52.4 & 49.8 & \cellcolor{lightorange!20} \textbf{53.5} & 49.5 & 50.3 & \cellcolor{lightorange!20} \textbf{53.9} \\
        
        \midrule
        
        \textcolor{Gray!50}{\textit{\textbf{Text / Doc / Chart}}} &  &  &  &  & \cellcolor{lightorange!20} &  &  & \cellcolor{lightorange!20} \\
        
        TextVQA~\cite{singh2019towards} & 71.5 & 68.1 & 72.0 & \textbf{72.6} & \cellcolor{lightorange!20} 71.9 & 70.2 & 72.3 & \cellcolor{lightorange!20} \textbf{73.6} \\
        
        ChartQA~\cite{masry2022chartqa} & 71.3 & 72.3 & 73.8 & 72.3 & \cellcolor{lightorange!20} \textbf{74.0} & 75.8 & 76.4 & \cellcolor{lightorange!20} \textbf{78.0} \\
        
        CharXiV$_{\text{dq}}$~\cite{wang2024charxiv} & 40.7 & 40.3 & 39.3 & 41.1 & \cellcolor{lightorange!20} \textbf{41.3} & 42.7 & \textbf{45.6} & \cellcolor{lightorange!20} 41.8 \\
        
        CharXiV$_{\text{rq}}$~\cite{wang2024charxiv} & 15.4 & 16.2 & \textbf{21.3} & 15.6 & \cellcolor{lightorange!20} 18.2 & 18.2 & \textbf{24.6} & \cellcolor{lightorange!20} 22.9 \\
        
        DocVQA~\cite{mathew2021docvqa} & 79.1 & 82.5 & \textbf{84.5} & 81.8 & \cellcolor{lightorange!20} 82.9 & 86.0 & \textbf{87.8} & \cellcolor{lightorange!20} 87.2 \\
        
        InfoVQA~\cite{mathew2022infographicvqa} & 39.3 & 40.4 & \textbf{44.2} & 42.2 & \cellcolor{lightorange!20} \textbf{44.2} & 45.9 & 48.2 & \cellcolor{lightorange!20} \textbf{48.7} \\
        
        OCRBench~\cite{liu2023hidden} & 566 & 542 & \textbf{608} & 590 & \cellcolor{lightorange!20} 558 & 647 & 581 & \cellcolor{lightorange!20} \textbf{659} \\
        
        SEEDBench-2-Plus~\cite{li2024seed} & 60.0 & 61.3 & 62.7 & 61.8 & \cellcolor{lightorange!20} \textbf{63.6} & 62.8 & 64.1 & \cellcolor{lightorange!20} \textbf{65.1} \\
        
        \textit{Avg} & \textit{54.2} & \textit{54.4} & \textbf{\textit{57.3}} & \textit{55.8} & \cellcolor{lightorange!20} \textit{56.5} & \textit{58.3} & \textit{59.6} & \cellcolor{lightorange!20} \textbf{\textit{60.4}} \\
        
        \midrule
        
        \textcolor{Gray!50}{\textit{\textbf{General VQA}}} &  &  &  &  & \cellcolor{lightorange!20} &  &  & \cellcolor{lightorange!20} \\
        
        BLINK~\cite{fu2024blink} & 44.2 & \textbf{45.4} & 44.1 & 44.3 & \cellcolor{lightorange!20} 44.0 & \textbf{43.8} & 43.2 & \cellcolor{lightorange!20} 43.4 \\

        MME~\cite{fu2023mme} & 1952 & 1983 & 1897 & 1962 & \cellcolor{lightorange!20} \textbf{1995} & 2029 & 2000 & \cellcolor{lightorange!20} \textbf{2031} \\
        
        MMBench$_{\text{en\_dev}}$~\cite{liu2023mmbench} & 75.2 & 74.1 & \textbf{76.4} & 75.9 & \cellcolor{lightorange!20} 76.0 & 73.2 & 73.9 & \cellcolor{lightorange!20} \textbf{76.5} \\
        
        MMBench$_{\text{cn\_dev}}$~\cite{liu2023mmbench} & 74.3 & 73.3 & \textbf{75.4} & 74.6 & \cellcolor{lightorange!20} 75.3 & 70.8 & 72.1 & \cellcolor{lightorange!20} \textbf{74.1} \\
        
        MMBench-v1.1$_{\text{en\_dev}}$~\cite{liu2023mmbench} & 73.0 & 72.0 & \textbf{74.6} & 73.9 & \cellcolor{lightorange!20} 73.6 & 70.5 & 72.1 & \cellcolor{lightorange!20} \textbf{74.5} \\
        
        MMBench-v1.1$_{\text{cn\_dev}}$~\cite{liu2023mmbench} & 72.8 & 71.4 & \textbf{74.1} & 73.1 & \cellcolor{lightorange!20} 73.5 & 68.1 & 70.0 & \cellcolor{lightorange!20} \textbf{73.2} \\
        
        MMMU-Pro$_{\text{standard}}$~\cite{yue2024mmmu} & 27.3 & 26.8 & \textbf{28.2} & 28.0 & \cellcolor{lightorange!20} 27.2 & 28.1 & 27.9 & \cellcolor{lightorange!20} \textbf{29.3} \\
        
        MMMU-Pro$_{\text{vision}}$~\cite{yue2024mmmu} & 12.3 & 12.9 & 12.1 & \textbf{13.0} & \cellcolor{lightorange!20} 12.3 & 13.5 & 14.2 & \cellcolor{lightorange!20} \textbf{14.6} \\
        
        MMStar~\cite{chen2024we} & 51.1 & 51.3 & 51.9 & 50.1 & \cellcolor{lightorange!20} \textbf{52.9} & 50.4 & 53.7 & \cellcolor{lightorange!20} \textbf{53.9} \\
        
        MMVet~\cite{yu2023mmvet} & 44.7 & 43.7 & 44.1 & 43.9 & \cellcolor{lightorange!20} \textbf{46.3} & 42.2 & 44.1 & \cellcolor{lightorange!20} \textbf{47.1} \\
        
        RealWorldQA~\cite{realworldqa} & 60.3 & 62.1 & 61.6 & 60.9 & \cellcolor{lightorange!20} \textbf{64.6} & 59.0 & 61.4 & \cellcolor{lightorange!20} \textbf{61.7} \\
        
        \textit{Avg} & \textit{55.0} & \textit{54.9} & \textit{55.5} & \textit{55.3} & \cellcolor{lightorange!20} \textbf{\textit{56.1}} & \textit{53.8} & \textit{54.9} & \cellcolor{lightorange!20} \textbf{\textit{55.5}} \\
        
        \bottomrule
    \end{tabular}
    \end{small}
\end{table}
\renewcommand{\arraystretch}{1.0}

\minisection{Experiment settings.} To compare with state-of-the-art vision encoders, we first pre-train the \model encoder at 1512 resolution using SigLIP2-SO400M~\cite{tschannen2025siglip} as the initialization, which we call \model-1.5K-SigLIP2. Then we evaluate \model-1.5K-SigLIP2 as well as each baseline vision encoder by training an MLLM on top of it using the same training recipe as in Section~\ref{sec:scaling_property} and testing it on various MLLM benchmarks. We do not compare the 4K-resolution PS3 here because we are not able to train MLLM with any baseline encoder at 4K resolution due to OOM issue and thus cannot build a fair comparison between them and PS3-4K-SigLIP2. We compare \model-1.5K-SigLIP2 with state-of-the-art vision encoders including SigLIP2-SO400M~\cite{tschannen2025siglip}, C-RADIOv2-L~\cite{Ranzinger_2024_CVPR}, PE$_{\text{Core}}$-L~\cite{bolya2025perception}, and AIMv2-L~\cite{fini2024multimodal}. Additionally, we follow the previous literature~\cite{bolya2025perception} and present PS3$_{\text{Lang}}$-SigLIP2-1.5K which is the \model model that is further aligned with LLMs by co-training with \vilamodel (Section~\ref{sec:sota_mllm}). Specifically, after training \vilamodel, we take out the vision encoder part as an independent vision encoder. We compare the LLM-aligned version of \model with other LLM-aligned vision encoders such as PE$_{\text{Lang}}$-L~\cite{bolya2025perception} and InternViT2.5-300M~\cite{chen2024expanding}. For a fair comparison, we train MLLMs with each vision encoder at roughly the same resolution around 1.5K to 1.7K. Since every vision encoder except for \model is pre-trained at low resolutions, we use \stwo to scale them to high resolution when training MLLMs. We evaluate on three types of benchmarks, including high-resolution VQA, VQA related to texts, documents, and charts, and general VQA.

\minisection{Results.} Table~\ref{tab:sota_vision_encoder} shows the full results. For vision encoders without LLM alignment, PE$_{\text{Core}}$ is the most competitive encoder between the baselines and \model-SigLIP2-1.5K is able to obtain better performance for high-resolution VQA and general VQA. Especially, \model improves the performance by 2\% - 4\% on high-res benchmarks such as 4KPro and HRBench, indicating the effectiveness of high-res pre-training of \model. Note that \model is worse than PE$_{\text{Core}}$ on MME-RealWorld, which is possibly because the benchmark requires capabilities such as counting that are beyond simple high-resolution perception which is the main pre-training objective of \model. For benchmarks with texts, documents, and charts, \model is competitive with the state-of-the-art encoder while achieving the best results on benchmarks such as ChartQA and CharXiV$_{\text{dq}}$, which is partially credited to the document images in the pre-training data. However, the \model pre-training data is not specifically designed for text recognition in natural images, as reflected by the lower performance on OCRBench and TextVQA, leaving huge room for improvement in the future. For general VQA, \model also achieves the best performance on average, showing significant improvements on benchmarks such as MMVet and RealWorldQA. On the other hand, PS3$_{\text{Lang}}$-SigLIP2-1.5K achieves significantly better results compared to other encoders with or without LLM alignment, rendering it the state-of-the-art vision encoder across the benchmarks we evaluate here. Compared to PE$_{\text{Lang}}$, PS3$_{\text{Lang}}$ dominates the performance on most of the benchmarks, including 8.0\% and 10.9\% improvements on highres benchmarks of 4KPro and HRBench$_{\text{4K}}$. It is also worth noting that if we compare the \model encoders with and without LLM alignment, the improvements are mostly on the text-related benchmarks, which implies the current MLLM training is still language-centric and fails to improve the vision-centric capabilities of vision encoders.

\finding{3}{Compared to previous vision encoders with low-resolution pre-training only, \model achieves state-of-the-art results on various MLLM benchmarks, especially the ones that require high-resolution. Aligning \model with LLMs further boosts its performance on text-related benchmarks.}

\subsection{Comparing \vilamodel to State-Of-The-Art MLLMs}
\label{sec:sota_mllm}

\begin{table*}[t]
\caption{\textbf{Comparing \vilamodel to state-of-the-art MLLMs on high-resolution benchmarks. Here \vilamodel-8B-4K refer to \textbf{\vilamodel-8B-\model-4K-SigLIP2}}. Bold and underlined results mean 1st and 2nd best performance for each task. \vilamodel-8B-4K achieves top-2 performance in 4 of the 6 benchmarks. $^\dagger$For most of the models, the resolution and number of tokens are dependent on the maximum number of tiles to use at test time, and we set the maximum number of tiles such that the number of tokens are similar to \vilamodel-8B-4K when it is selecting 18\% patches for a fair comparison.
}
    \label{tab:sota_highres}
    \centering
    \begin{small}
    \begin{tabular}{lllcccccc}
        \toprule
         & Res.$^\dagger$ & \#Token$^\dagger$ & 4KPro & HRBench$_{\text{4K}}$ & HRBench$_{\text{8K}}$ & MME-RW & V$^\ast$$_{\text{att}}$ & V$^\ast$$_{\text{pos}}$\\

        \midrule
        NVILA-8B~\cite{liu2024nvila} & 1552 & 3072 & 58.1 & 59.0 & 55.5 & 52.0 & 68.7 & 65.8 \\
        PLM-8B~\cite{cho2025perceptionlm} & 1735 & 4096 & 59.7 & \underline{\textbf{71.3}} & 61.1 & \underline{\textbf{63.1}} & 72.2 & 68.4 \\
        InternVL2.5-8B~\cite{chen2024expanding} & 1735 & 4096 & 66.1 & 65.5 & 57.4 & 58.9 & 75.8 & \underline{\textbf{76.5}} \\
        InternVL3-8B~\cite{zhu2025internvl3} & 1735 & 4096 & 56.5 & 67.5 & 62.5 & \underline{60.7} & 72.2 & \underline{75.0} \\
        Qwen2-VL-7B~\cite{wang2024qwen2} & 1735 & 3840 & \underline{69.4} & 66.9 & 62.1 & 58.1 & \underline{76.5} & 67.1 \\ 
        Qwen2.5-VL-7B~\cite{bai2025qwen2} & 1735 & 3840 & 66.1 & 65.6 & \underline{64.1} & 57.1 & \underline{\textbf{78.3}} & 71.1 \\
        \rowcolor{lightorange!20} \vilamodel-8B-4K & 3780 & 4036 & \underline{\textbf{71.0}} & \underline{67.8} & \underline{\textbf{65.3}} & 50.9 & \underline{76.5} & 71.1 \\
        \bottomrule
    \end{tabular}
    \end{small}
\end{table*}

\begin{table*}[t]
\caption{\textbf{Comparing \vilamodel to state-of-the-art MLLMs on other common benchmarks.} Here \vilamodel-8B-1.5K and \vilamodel-8B-4K refer to \textbf{\vilamodel-8B-\model-1.5K-SigLIP2} and \textbf{\vilamodel-8B-\model-4K-SigLIP2}, respectively. \textit{Res.} is the maximum resolution each model supports. Some models (\eg, Qwen2-VL, InternVL2) can accept input images of different aspect ratios, for which the resolution is calculated as square root of the maximum number of pixels the model can take in. \textit{Select} is the high-res patch selection ratio of \model at test time. \textit{\#Token} is the total number of visual tokens fed into LLM under the maximum input resolution. \vilamodel-8B-4K achieves state-of-the-art performance on high-resolution benchmarks such as V$^\ast$Bench and 4KPro.
}
    \label{tab:sota}
    \centering
    \begin{scriptsize}
    \begin{tabular}{lp{0.03\textwidth}p{0.03\textwidth}p{0.05\textwidth}|cccccccccccc}
        \toprule
         & Res. & Select & \#Token & \multirowver{ChartQA \\ (test)} & \multirowver{DocVQA \\ (test)} & \multirowver{InfoVQA \\ (test)} & \multirowver{MathVista \\ (testmini)} & \multirowver{MMBench \\ (en-dev)} & \multirowver{MMMU-Pro \\ (standard)} & \multirowver{OCRBench \\ (test)} & \multirowver{V$^\ast$Bench \\ (test)} & \multirowver{RealWorldQA \\ (test)} & \multirowver{TextVQA \\ (val)} & \multirowver{4KPro \\ (test)} \\
        \midrule
        \textit{Proprietary} \\
        GPT-4o~\cite{hurst2024gpt} & - & - & - & 85.7 & 92.8 & - & 63.8 & - & 54.0 & 736 & 53.7 & 58.6 & - & 59.7 \\
        Claude 3.5 Sonnet~\cite{claude35sonnet} & - & - & - & 90.8 & 95.2 & 49.7 & 67.7 & - & 55.0 & 788 & 23.0 & 59.9 & - & 29.0 \\
        Gemini-1.5-Pro~\cite{team2024gemini} & - & - & - & 87.2 & 93.1 & 81.0 & 63.9 & - & 49.4 & 754 & 60.3 & 70.4 & 78.7 & 59.7 \\
        \midrule
        \textit{Open-source} \\

        VILA-1.5-8B~\cite{lin2023vila} & 336 & - & 576 & 52.7 & 40.6 & 25.9 & 36.7 & 68.9 & - & - & - & 52.7 & 68.5 & 33.9 \\
        Cambrian-1-8B~\cite{tong2024cambrian} & 1024 & - & - & 73.3 & 77.8 & - & 49.0 & 75.9 & - & 624 & 59.2 & 64.2 & 71.7 & 50.0 \\
        MM1.5-7B~\cite{zhang2024mm1} & 2016 & - & 1440 & 78.6 & 88.1 & 59.5 & 47.6 & - & - & 635 & - & 62.5 & 76.5 & - \\
        NVILA-8B~\cite{liu2024nvila} & 1552 & - & 3072 & \textbf{86.1} & 93.7 & 70.7 & 65.4 & 87.6 & 33.6 & 794 & 67.2 & 66.4 & 80.1 & 58.1 \\
        LLaVA-OV-7B~\cite{li2024llava} & 2304 & - & 7252 & 80.0 & 87.5 & 68.8 & 63.2 & 80.8 & 29.5 & - & 69.2 & 66.3 & - & 67.7 \\
        IXC2-4KHD~\cite{dong2024internlm} & 2479 & - & 7920 & 81.0 & 90.0 & 68.6 & 57.8 & 80.2 & - & 675 & - & - & 77.2 & 42.8 \\
        IXC-2.5-7B~\cite{zhang2024internlm} & 2743 & - & 10000 & 82.2 & 90.9 & 70.0 & 59.6 & 82.2 & - & 690 & 45.6 & 67.8 & 78.2 & 32.3 \\
        InternVL2-8B~\cite{internvl2} & 2833 & - & 10496 & 83.3 & 91.6 & 74.8 & 58.3 & 81.7 & 32.5 & 794 & 65.8 & 64.4 & 77.4 & 58.1 \\
        Qwen2-VL-7B~\cite{wang2024qwen2} & 3584 & - & 16384 & 83.0 & \textbf{94.5} & \textbf{76.5} & 58.2 & - & - & \textbf{866} & 71.0 & 70.1 & \textbf{84.3} & \textbf{71.0} \\

        \midrule
        \rowcolor{lightorange!20} & 1512 & 33\% & 1411 & 84.4 & 89.5 & 62.3 & 66.2 & 92.0 & 35.0 & 795 & 66.5 & 69.3 & 79.5 & 59.7 \\
        \rowcolor{lightorange!20} & 1512 & 67\% & 2626 & 85.8 & 92.8 & 68.0 & 65.8 & 92.3 & \textbf{35.2} & 817 & 68.0 & 69.7 & 80.3 & 67.7 \\
        \rowcolor{lightorange!20} \multirow{-3}{*}{\vilamodel-8B-1.5K} & 1512 & 100\% & 3841 & 85.4 & 93.1 & 70.0 & \textbf{66.6} & 92.4 & 35.0 & 819 & 68.5 & \textbf{70.8} & 80.4 & 67.7 \\
        
        \midrule
        \rowcolor{lightorange!20} & 3780 & 6\% & 1476 & 83.6 & 89.0 & 59.7 & 65.8 & 91.7 & 34.9 & 791 & 66.5 & 68.2 & 79.9 & 56.5 \\
        \rowcolor{lightorange!20} & 3780 & 12\% & 2756 & 84.7 & 92.3 & 66.3 & 66.2 & 92.4 & \textbf{35.2} & 794 & 69.7 & 69.5 & 80.4 & 59.7 \\
        \rowcolor{lightorange!20} \multirow{-3}{*}{\vilamodel-8B-4K} & 3780 & 18\% & 4036 & 84.7 & 92.7 & 68.2 & \textbf{66.6} & \textbf{92.6} & 34.5 & 802 & \textbf{73.3} & 69.9 & 80.5 & \textbf{71.0} \\

        \bottomrule
    \end{tabular}
    \end{scriptsize}
\end{table*}

\minisection{Experiment settings.} To compare with state-of-the-art MLLMs, we train \vilamodel using the full recipe in NVILA-8B~\cite{liu2024nvila} (except that we use less data in stage 2 as described in Appendix~\ref{appendix_sec:mllm_training_setting}) and compare its performance with state-of-the-art MLLMs. We train \vilamodel with two \model encoders, \model-1.5K-SigLIP2 and \model-4K-SigLIP2, and denote them by \vilamodel-8B-\model-1.5K-SigLIP2 and \vilamodel-8B-\model-4K-SigLIP2, respectively. We evaluate on several high-resolution VQA benchmarks (4KPro, HRBench~\cite{wang2025divide}, MME-RealWorld~\cite{zhang2024mme}, and V$^\ast$~\cite{wu2024v}) and other general VQA benchmarks (TextVQA~\cite{singh2019towards}, ChartQA~\cite{masry2022chartqa}, DocVQA~\cite{mathew2021docvqa}, InfoVQA~\cite{mathew2022infographicvqa}, MathVista~\cite{lu2023mathvista}, MMBench~\cite{liu2023mmbench}, MMMU-Pro~\cite{yue2024mmmu},  OCRBench~\cite{liu2023hidden}, and RealWorldQA~\cite{realworldqa}). For some benchmarks, we balance the patch selection ratio across different scales such that it reaches the optimal performance at test time. Please see Appendix~\ref{appendix_sec:mllm_training_setting} for more detailed training recipe and evaluation setting. 

\minisection{Results.} We first show the results on high-res benchmarks in Table~\ref{tab:sota_highres}. We compare with state-of-the-art MLLMs under approximately the same number of vision tokens. We can see \vilamodel-8B-4K can process higher resolution with similar number of tokens, thanks to the top-down patch selection mechanism. As a result, \vilamodel achieves the best results on 4KPro and HRBench$_{\text{8K}}$ with improvements of 4.9\% and 2.2\% respectively on 4KPro and HRBench$_{\text{8K}}$ over Qwen2.5-VL-7B~\cite{bai2025qwen2}. On HRBench$_{\text{4K}}$ and V$^\ast_\text{att}$, \vilamodel also achieves second-to-best results. Notably, \vilamodel achieves 7.3\% improvement on average over NVILA-8B, which showcases the advantage of high-res vision pre-training considering it is the only difference between the two models. The only benchmark where \vilamodel fails to perform well is MME-RealWorld, which is probably because the benchmark requires more complicated capabilities such as counting which is beyond the scope of \model pre-training which is focused on basic capabilities of high-res perception and recognition.

We show the results on other general benchmarks in Table~\ref{tab:sota}, \vilamodel shows competitive performance compared to state-of-the-art MLLMs such as NVILA and Qwen2-VL and achieves the best results in 6 out of 11 benchmarks. Specifically, \vilamodel-1.5K achieves the best results on benchmarks that have the MRR around 512-1K including MathVista and MMMU-Pro, and \vilamodel-4K obtains state-of-the-art performance on benchmarks that require more detailed understanding (MRR between 1K and 4K) such as V$^\ast$Bench and 4KPro, surpassing NVILA despite using the same recipe and less data, showing the effect of \model pre-training. \vilamodel-4K also outperforms all proprietary MLLMs on these high-MRR benchmarks, for example, improving by 13.0\% and 11.3\% over Gemini-1.5-Pro on V$^\ast$Bench and 4KPro. Note that this is achieved by selecting only 18\% of the high-res patches, showing both the efficacy and efficiency of our model. On the other hand, \vilamodel has slightly worse results than NVILA and Qwen2-VL on OCR-related benchmarks including ChartQA, DocVQA, TextVQA, InfoVQA, and OCRBench. This is probably because the pre-training data of \model is not specifically optimized for OCR and text recognition and adding such data in pre-training can potentially solve the problem. We further show that \model can achieve even higher efficiency with only minor performance degradation on most benchmarks. Specifically, by selecting only 6\% patches for \model, it maintains competitive performance of \vilamodel-4K while only using 1476 tokens for a maximum resolution of 4K which is less than 1/10 of \#token of Qwen2-VL. We can see the performance stays similar compared to 18\% patch selection for benchmarks such as TextVQA and RealWorldQA, and only has minor drops for CharQA and V$^\ast$Bench. Tasks like DocVQA require denser visual information and thus have larger performance drop.

\subsection{Comparing \model to State-Of-The-Art Token Pruning Methods}

\begin{table}
\caption{\textbf{Comparing \model to state-of-the-art token pruning methods.} \model has consistently lower ViT latency and achieve better performances than previous methods. \model is also the only method that can handle 4K resolution images.}
    \label{tab:sota_token_pruning}
    \vspace{-0.5em}
    \centering
    \begin{small}
    \begin{tabular}{llcccccccccc}
        \toprule
        Method & \makecell[l]{Select \\ (Test)} & \makecell{ViT \\ Latency} & \makecell{LLM \\ Latency} & \makecell{Text \\ VQA} & \makecell{Chart \\ QA} & \makecell{Doc \\ VQA} & \makecell{Info \\ VQA} & \makecell{OCR \\ Bench} & \makecell{V$^\ast$ \\ Bench}  & \makecell{Real \\ World}  & \textit{Avg} \\ 
        \midrule\midrule
        \rowcolor{Gray!20} \multicolumn{12}{c}{\textit{\textcolor{Gray}{\textbf{1512 Resolution}}}} \\
        Full & 100\% & 0.286s & 0.375s & 78.6 & 84.1 & 92.2 & 68.1 & 787 & 67.9 & 69.8 & \textit{77.1} \\
        \midrule
        ToMe~\cite{bolya2022token}  & 50\% & 0.286s & 0.260s & 74.1 & 70.2 & 59.7 & 47.3 & 622 & 66.8 & 67.2 & \textit{63.9} \\
        FastV~\cite{chen2025image} & 50\% & 0.286s & 0.264s & \textbf{78.2} & 81.2 & \textbf{90.0} & 60.4 & 769 & 66.2 & 69.0 & \textit{74.6} \\
        VisionZip~\cite{yang2024visionzip} & 50\% & 0.286s & 0.260s & 75.2 & 77.2 & 79.8 & 55.7 & 722 & 64.0 & 67.1 & \textit{70.2} \\
        \rowcolor{lightorange!20} \model & 50\% & \textbf{0.167s} & 0.260s & 77.7 & \textbf{83.4} & 89.8 & \textbf{60.8} & \textbf{774} & \textbf{67.9} & \textbf{69.1} & \textit{\textbf{75.2}} \\
        \midrule
        ToMe~\cite{bolya2022token}  & 25\% & 0.286s & 0.180s & 72.5 & 65.5 & 51.7 & 42.8 & 61.1 & 62.2 & 63.4 & \textit{59.9} \\
        FastV~\cite{chen2025image} & 25\% & 0.286s & 0.185s & 76.1 & 66.3 & 78.1 & 49.5 & 651 & 64.6 & 65.2 & \textit{66.6} \\
        VisionZip~\cite{yang2024visionzip} & 25\% & 0.286s & 0.180s & 74.6 & 76.0 & 72.8 & 51.5 & 694 & 62.7 & 64.6 & \textit{67.4} \\
        \rowcolor{lightorange!20} \model & 25\% & \textbf{0.096s} & 0.180s & \textbf{76.8} & \textbf{80.4} & \textbf{84.4} & \textbf{54.6} & \textbf{738} & \textbf{65.7} & \textbf{67.8} & \textit{\textbf{71.9}} \\

        \midrule\midrule
        \rowcolor{Gray!20} \multicolumn{12}{c}{\textit{\textcolor{Gray}{\textbf{3780 Resolution}}}} \\
        Full & 100\% & 1.812s & OOM & - & - & - & - & - & - & - & - \\
        \midrule
        ToMe~\cite{bolya2022token}  & 20\% & 1.812s & OOM & - & - & - & - & - & - & - & -  \\
        FastV~\cite{chen2025image} & 20\% & 1.812s & OOM & - & - & - & - & - & - & - & - \\
        VisionZip~\cite{yang2024visionzip} & 20\% & 1.812s & OOM & - & - & - & - & - & - & - & - \\
        \rowcolor{lightorange!20} \model & 20\% & \textbf{0.417s} & 0.383s & 77.8 & 83.9 & 91.6 & 65.0 & 773 & 72.8 & 70.1 & \textit{76.9} \\
        \bottomrule
    \end{tabular}
    \end{small}
\end{table}

\minisection{Experiment settings.} We compare the efficiency of \model on high-res images with state-of-the-art token pruning methods for MLLMs including ToMe~\cite{bolya2022token}, FastV~\cite{chen2025image}, and VisionZip~\cite{yang2024visionzip}.  We test the models in both 1512 and 3780 resolution and try different token selection ratio for each resolution. All the baselines use \model as the vision encoder for a fair comparison but instead of using the \model patch selection module, they use \model to process the whole image and prune the tokens in their own way. The \model encoder is pre-trained on top of SigLIP-SO400M as in Section~\ref{sec:scaling_property}.

\minisection{Results.} Full results are shown in Table~\ref{tab:sota_token_pruning}. First of all, we can see that \model achieves similar efficiency on the LLM backbone when pruning the same number of tokens, while significantly reduces the ViT latency (\eg, 0.286s $\rightarrow$ 0.096s when selecting 25\% patches at 1512 resolution). This is because \model reduces the tokens in both ViT and LLM while other methods only prune the tokens in LLM. This advantage is especially important for 4K-resolution images where the ViT latency dominates the LLM latency. For the same reason, the previous methods all run out of memory on 4K resolution images while \model is able to process 4K resolution efficiently. In the meantime, \model achieves superior results over all previous methods under the same selection ratio. This is due to several possible reasons. For example, the previous methods are heuristic-based, either merging visual tokens based on their similarities or pruning tokens based on the attention score in the vision encoder or LLM backbone, while \model adopts a learning-based approach to select important or relevant visual tokens, which is more accurate than heuristic-based methods given enough data. On the other hand, unlike previous methods such as ToMe and VisionZip that are prompt-agnostic, \model is able to select patches based on the user prompt which can be more accurate.

\finding{4}{\model's learning-based top-down patch selection achieves better performance and efficiency than heuristic-based token pruning methods.}

\subsection{Generalizability of \model Pre-Training to State-Of-The-Art Vision Encoders}

\begin{table}
\caption{\textbf{Generalizability of \model pre-training to state-of-the-art vision encoders.} \textit{\#Param of ViT} is the number of parameters of the ViT backbone. \textit{Max Res.} is the maximum resolution each model processes in MLLM. \textit{Max \#Tok} is the maximum number of vision tokens in MLLM. All the \model models select all the high-res patches.}
    \label{tab:sota_vision_encoder}
    \vspace{-0.5em}
    \centering
    \begin{small}
    \begin{tabular}{p{0.19\textwidth}p{0.055\textwidth}p{0.045\textwidth}p{0.045\textwidth}cccccccc}
        \toprule
        \makecell[l]{Vision \\ Encoder} & \makecell[l]{\#Param \\ of ViT} & \makecell[l]{Max \\ Res.} & \makecell[l]{Max \\ \#Tok} & \makecell{Text \\ VQA} & \makecell{Chart \\ QA} & \makecell{Doc \\ VQA} & \makecell{Info \\ VQA} & \makecell{OCR \\ Bench} & \makecell{V$^\ast$ \\ Bench}  & \makecell{Real \\ World}  & \textit{Avg} \\ 
        \midrule
        SigLIP-SO400M~\cite{zhai2023sigmoid} & 400M & 378 & 196 & 62.3 & 56.6 & 51.9 & 30.7 & 387 & 51.8 & 57.1 & \textit{49.9} \\
        \ \ + AnyRes~\cite{liu2024llavanext} & 400M & 1512 & 3332 &  67.4 & 58.4 & 67.9 & 34.1 & 468 & 60.2 & 59.0 & \textit{56.3} \\
        \ \ + \stwo~\cite{shi2025we} & 400M & 1512 & 2916 & 66.1 & 71.0 & 78.3 & 41.1 & 526 & 55.2 & 61.0 & \textit{60.8} \\
        \rowcolor{lightorange!20} \model-SigLIP-SO400M & 400M & 1512 & 3841 & \textbf{69.3} & \textbf{71.1} & \textbf{79.4} & \textbf{41.3} & \textbf{534} & \textbf{64.0} & \textbf{63.8} & \textbf{\textit{63.2}} \\
        \midrule
        C-RADIOv2-L~\cite{Ranzinger_2024_CVPR}  & 320M & 384 & 144 & 65.0 & 58.8 & 53.1 & 30.9 & 405 & 51.5 & 57.5 & \textit{51.0} \\
        \ \ + AnyRes~\cite{liu2024llavanext} & 320M & 1536 & 2448 & 68.1 & 62.8 & 70.0 & 35.8 & 497 & 65.9 & 62.8 & \textit{59.3} \\
        \ \ + \stwo~\cite{shi2025we} & 320M & 1536 & 2304 & 68.1 & 72.3 & 82.5 & 40.4 & 542 & 59.7 & 62.1 & \textit{62.8} \\
        \rowcolor{lightorange!20} \model-C-RADIOv2-L & 320M & 1536 & 3024 & \textbf{68.4} & \textbf{72.6} & \textbf{83.2} & \textbf{43.4} & \textbf{569} & \textbf{68.2} & \textbf{61.5} & \textbf{\textit{64.9}}  \\
        \bottomrule
    \end{tabular}
    \end{small}
\end{table}

We verify the generalizability of \model pre-training pipeline by pre-training \model on top of several state-of-the-art vision encoders and compare with the same vision encoders without high-res pre-training. Specifically, we take the state-of-the-art language-aligned vision model, SigLIP-SO400M~\cite{zhai2023sigmoid}, and agglomerative vision model, C-RADIO-v2-L~\cite{Ranzinger_2024_CVPR}, as the baselines. We take their pre-trained model and continue pre-training on high-res images using \model pre-training pipeline to get \model-SigLIP-SO400M and \model-C-RADIO-v2-L. We compare them with the original models as well as their AnyRes~\cite{liu2024llavanext} and \stwo~\cite{shi2025we} variants. Table~\ref{tab:sota_vision_encoder} shows the results. We can see that for both the base models, the \stwo variant achieves the best performance compared to AnyRes variant as well as the original version, while \model encoders pre-trained on top of each base model are able to achieve better results, showing that \model is a general high-res pre-training pipeline that can be applied to any base vision encoder. We also observe that C-RADIO baseline consistently outperforms SigLIP baseline, and \model-C-RADIO-v2-L is able to inherit that advantage and outperforms \model-SigLIP.

\section{Ablations and Analysis}
\label{sec:ablation}

\subsection{Key Designs in Pre-Training Algorithm and Model Architecture}
\label{sec:ablation_model_algorithm}

We conduct ablation studies on the pre-training algorithm designs (Section~\ref{sec:pretrain_algorithm}), \model model designs (Section~\ref{sec:pretrain_model_design}), and MLLM model design (Section~\ref{sec:ps3_for_vila}). We follow the same experimental setting as in Section \ref{sec:scaling_property} and use \model with max resolution of 1512 and 100\% patch selection as the baseline. For the ablation of each design, we report the improvement of the baseline model on the average accuracy of the seven benchmarks compared to the model without the design. Results are shown in Table~\ref{tab:ablation_pretrain_and_model}. Overall all the designs are helpful. Among the pre-training algorithm designs, we can see that it is crucial to select the patches in the ground-truth boxes during patch selection and pool only their corresponding tokens before calculating the contrastive loss. This is because otherwise the model will select irrelevant patches and contrast them with the local caption, leading to noisy pre-training and degrading the performance by over 8$\%$. We also find avoiding contrast between local regions in the same image improves the performance by 5.3\% which aligns with the observation in previous work~\cite{chen2024contrastive}. The model architecture designs also help with the performance. Notably, the low-res KV cache allows the model to see the global context while extracting the local high-res features, which significantly improves the performance by 8.8\%. Additionally, extracting features at multiple scales (\eg, 756 and 1512) performs better than only extracting features at the largest scale (\eg, 1512 only) as in AnyRes. Adding scale-aware positional embeddings in \model and additional vision positional embeddings when feeding \model features to LLM also improve the performance by a small margin.

\finding{5}{High-resolution pre-training with localized contrastive learning requires several key algorithm designs, such as contrasting within the ground-truth selection boxes and avoiding contrasting between different regions in the same image, and key model designs including low-res KV cache that makes high-res features aware of the global context.}

\subsection{Top-Down and Bottom-Up Patch Selection Matters}
\label{sec:ablation_patch_selection}

We compare the performance of using random, bottom-up, and top-down patch selection, as shown in Table~\ref{tab:ablation_top_down_selection}. We report benchmark accuracy as well as the patch recall rate which evaluates how many patches out of the ground-truth region are selected at test time. This is evaluated on a test set split from the data used to train patch selection in MLLM. We can see that both bottom-up and top-down selection significantly improves the recall and the accuracy over random selection. For example, when selecting 44\% patches, bottom-up selection improves the recall rate by 43.7\% and the accuracy by 7.4\% over random selection. Top-down selection further improves the recall rate by 3.8\% and the accuracy by 1.3\%. Interestingly, patch selection not only affects test time but training as well. When we train MLLM with 44\% patches, even if we select all patches at test time, top-down selection is still better than the other two. The reason is likely that top-down selection provides more informative visual context for MLLM during training, which leads to less noisy training dynamics. Note that this model has performance comparable to training with 100\% patches, which means that training with top-down patch selection improves training efficiency without hurting the performance too much.

\finding{6}{Top-down and bottom-up patch selection provides more relevant visual information to MLLM, which does not only improve performance at test time but also improve the model optimization during training.}

\begin{table}[t]
\centering
\begin{minipage}{0.48\textwidth}
\caption{\textbf{Ablation of PS3 pre-training, model, and MLLM designs.} $\Delta$ is the change of the average performance on the seven benchmarks after adding the design.}
\label{tab:ablation_pretrain_and_model}
\vspace{-0.5em}
\begin{small}
\begin{tabular}{lc}
\toprule
   Training and Model Design Choices  & $\Delta$   \\
\midrule
\textit{Pre-training algorithm designs \ \textcolor{Gray}{(Section~\ref{sec:pretrain_algorithm})}}  \\
   \ \ - using ground truth selection score  & +5.1 \\
   \ \ - pooling only tokens in ground-truth boxes  & +3.7 \\
   \ \ - mixing global and local contrast  & +1.0 \\
   \ \ - w/o intra-image contrast  & +5.3 \\
\textit{\model model designs \ \textcolor{Gray}{(Section~\ref{sec:pretrain_model_design})}} \\
   \ \ - multi-scale feature extraction & +1.0 \\
   \ \ - scale-aware pos. emb. & +0.8 \\
   \ \ - low-res KV cache & +8.8 \\
\textit{MLLM model design \ \textcolor{Gray}{(Section~\ref{sec:ps3_for_vila})}}  \\
   \ \ - additional vision pos. emb. & +0.8 \\
\bottomrule
\end{tabular}
\end{small}
\end{minipage}
\hfill
\begin{minipage}{0.485\textwidth}
\caption{\textbf{Ablation of top-down and bottom-up patch selection.} \textit{Select (Train)} and \textit{Select (Test)} are the percentage of high-res patches \model selects at training and test time. \textit{Recall} is the recall rate of how many patches in the ground-truth regions are selected at test time.}
\label{tab:ablation_top_down_selection}
\vspace{-0.5em}
\begin{small}
\begin{tabular}{lcccc}
\toprule
   \makecell[l]{Patch \\ Selection} & \makecell[c]{Select \\ (Train)} & \makecell[c]{Select \\ (Test)} & \makecell[c]{Recall \\ (Test)} & Avg. Acc.   \\
\midrule
   Random & 44\% & 44\% & 43.7\% & 52.3 \\
   Bottom-up & 44\% & 44\% & 87.4\% & 59.7 \textcolor{nvidiagreen}{(+7.4)} \\
   Top-down & 44\% & 44\% & 91.2\% & 61.0 \textcolor{nvidiagreen}{(+8.7)} \\
\midrule
   Random & 44\% & 100\% & 100\% & 56.5 \\
   Bottom-up & 44\% & 100\% & 100\% & 61.1 \textcolor{nvidiagreen}{(+4.6)} \\
   Top-down & 44\% & 100\% & 100\% & 61.9 \textcolor{nvidiagreen}{(+5.4)} \\
\midrule
   Top-down & 100\% & 100\% & 100\% & 63.2 \\
\bottomrule
\end{tabular}
\end{small}
\end{minipage}
\end{table}

\subsection{Trade-Off Between Different Image Scales}
\label{sec:ablation_trafe_off_image_scales}

\begin{figure}[t]
    \centering
    \includegraphics[width=1\linewidth]{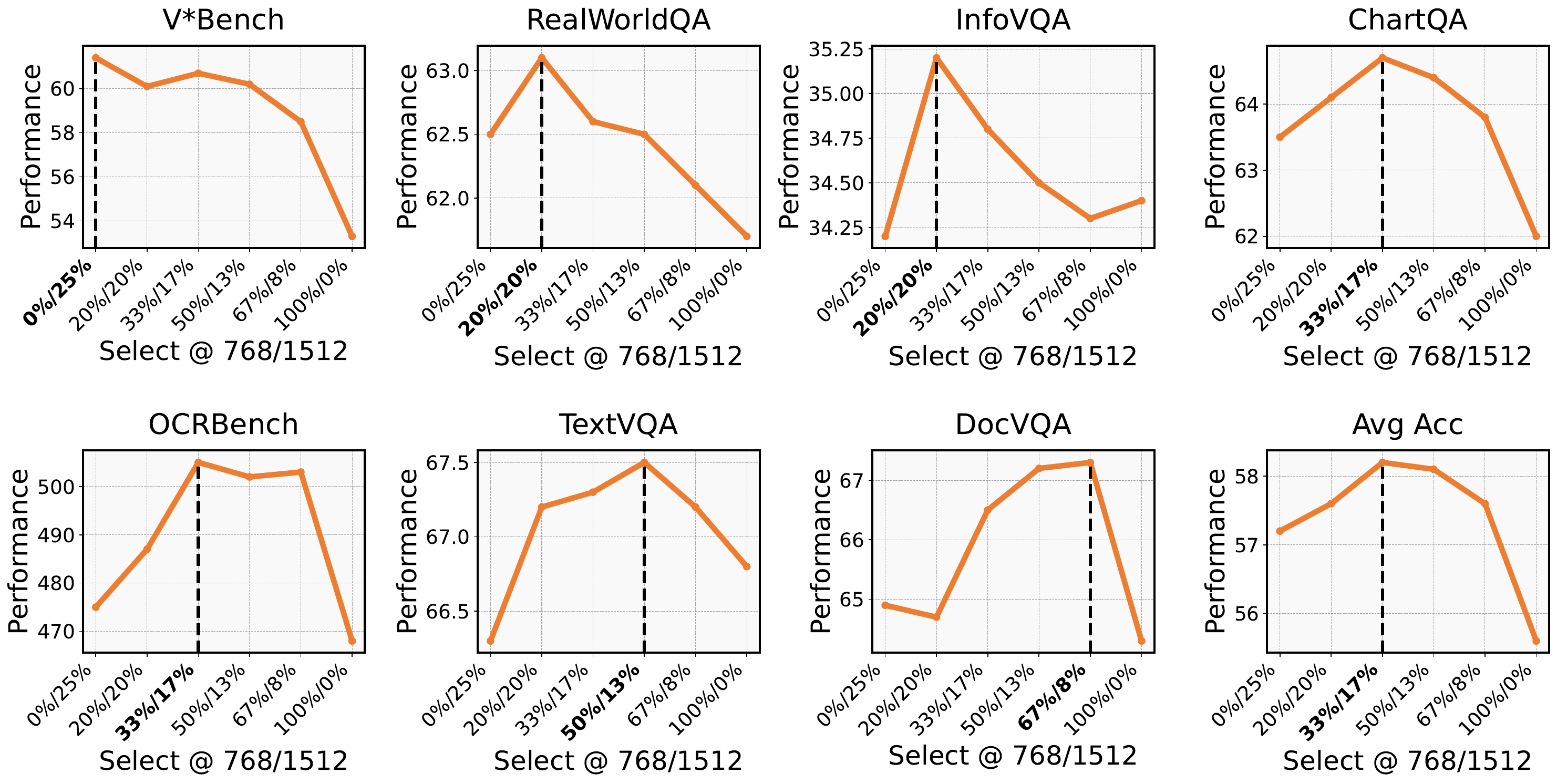}
    \caption{\textbf{Trade-off between image scales for different benchmarks.} \textit{Select @ 756/1512} are the percentage of selected patches at 756 and 1512 scales at test time, respectively. \model can flexibly adjust token selection ratios at different image scales to achieve the best performance for different downstream tasks. }
    \label{fig:ablation_selection_ratio}
\end{figure}

We find the optimal balance of patch selection between image scales varies for different tasks (Figure~\ref{fig:ablation_selection_ratio}). For example, when using in total 729 (20\%) high-res tokens, V$^\ast$Bench performance peaks when selecting no patches (0\%) at 756 scale and only patches (25\%) at 1512 scale, while it performs the best on DocVQA when selecting 67\% 756-scale patches and only 8\% 1512-scale patches. Note that selecting more patches at 756 scale covers more regions of the image because one 756-scale patch represents larger area in the original image than a 1512-scale patch, but lose more details at the 1512 scale. This indicates V$^\ast$Bench needs smaller coverage of image regions but requires more high-res information at 1512 resolution to perform well. DocVQA, on the other hand, needs more coverage of the image, which is probably because one usually needs to read the whole document to answer the questions. We also observe degraded performance on all benchmarks when selecting only tokens at 756 scale, which is because not all tokens at 756 scales are relevant to the question while the relevant information from 1512 scale is completely lost in this way.

\finding{7}{The optimal balance of patch selection between different image scales varies for different tasks. Selecting from smaller scales covers more image regions, which is better for tasks with denser visual information. Selecting from larger scales covers less region but captures more detailed information, which is favored by tasks that only require localized understanding. }

\subsection{SFT Data for High-Resolution Feature Alignment}
\label{sec:ablation_data}

We study the effect of high-resolution SFT data for MLLMs using \model as the vision encoder. Commonly-used image QA data normally contains only low-res images or lacks questions about details in high-res images, from which it is hard for MLLM to learn to utilize the high-res vision features. We hypothesize the high-res QA data in Section \ref{sec:train_vila_w_ps3}, though generated in a naive way, can help align the high-res visual features to the text space of LLM, thus improving the high-res perception capability. We conduct an ablation study in Table \ref{tab:ablation_highres_data}. We find that the high-res SFT data in general helps with the performance on resolution-sensitive benchmarks. The improvements are especially significant on natural images, \eg, a 3.1\% improvement on V$^\ast$Bench. On the other hand, the performance on ChartQA and DocVQA does not change, which is probably because the current high-res SFT data does not contain document images.

\begin{table}
\caption{\textbf{Ablation of MLLM SFT data for high-resolution feature alignment.} The high-resolution SFT data (\textit{HR Data}) generally improves high-resolution perception, especially on natural images.}
    \label{tab:ablation_highres_data}
    \vspace{-0.5em}
    \centering
    \begin{small}
    \begin{tabular}{lllccccccccc}
        \toprule
        \makecell[l]{Vision \\ Encoder} & \makecell[l]{Max \\ Res.} & \makecell[l]{HR \\ Data} & \makecell{Text \\ VQA} & \makecell{Chart \\ QA} & \makecell{Doc \\ VQA} & \makecell{Info \\ VQA} & \makecell{OCR \\ Bench} & \makecell{V$^\ast$ \\ Bench}  & \makecell{Real \\ World}  & \textit{Avg} \\ 
        \midrule
        \model & 1512 & \xmark & 68.8 & 71.2 & 79.6 & 39.6 & 535 & 60.9 & 62.9 & \textit{62.4} \\
        \model & 1512 & \cmark & 69.3 & 71.1 & 79.4 & 41.3 & 534 & 64.0 & 63.8 & \textit{63.2} \\
         &  &  & \textcolor{nvidiagreen}{(+0.5)} & (-0.1) & (-0.2) & \textcolor{nvidiagreen}{(+1.7)} & (-0.1) & \textcolor{nvidiagreen}{(+3.1)} & \textcolor{nvidiagreen}{(+0.9)} & \textcolor{nvidiagreen}{\textit{(+0.8)}} \\
        \bottomrule
    \end{tabular}
    \end{small}
\end{table}

\subsection{Visualization of \model Visual Representations}
\label{sec:pca}

\begin{figure}[t]
    \centering
    \includegraphics[width=1\linewidth]{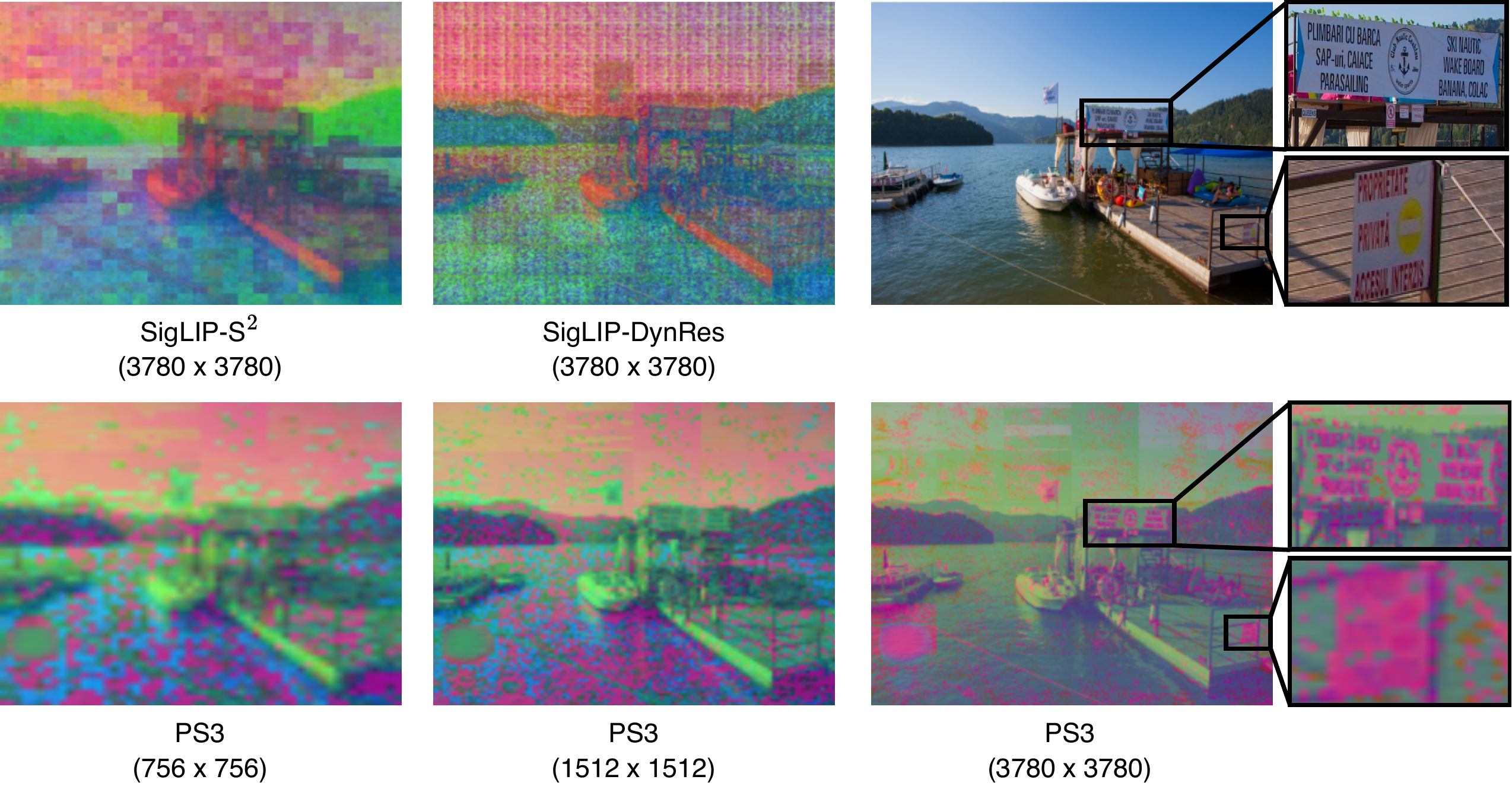}
    \caption{\textbf{PCA visualization of visual features.} The baselines, \stwo and AnyRes, have either noisy or blurry features at 4K resolution, while \model shows extremely fine-grained features that highlight details such as small texts on the banners.}
    \label{fig:pca}
\end{figure}

We visualize the visual features in \model using principal component analysis (PCA) and compare with the baselines of \stwo and AnyRes that does not pre-train vision models at high resolutions. Results are shown in Figure~\ref{fig:pca}. We visualize the features of SigLIP-\stwo and SigLIP-AnyRes at 3780$\times$3780 resolution and the features from three different scales (756$\times$756, 1512$\times$1512, and 3780$\times$3780) for \model. We can see the feature map of SigLIP-\stwo is blurry compared to other models because \stwo concatenates both low-res and high-res features and the PCA visualization is partially dominated by the low-res components. The AnyRes feature map, which is equivalent to \stwo but only with the high-res features, shows noisy patterns. One possible reason is AnyRes splits the high-res image into small tiles and individually processes each, which means the features lack global context and are inconsistent when running PCA on the whole feature map. On the other hand, \model shows sharp and fine-grained features at 4K resolution, \eg, at the edges of objects. We also observe \model features from different scales tend to group pixels at different scales. For example, the features at 756 and 1512 scales group large-scale objects or things such as the dock and the sky, while the features at 3780 resolution usually group small-scale patterns such as the texts on the banners.

\section{Related Work}

\minisection{Scaling up vision pre-training.} 
The enormous efforts from the past few years in both supervised pre-training~\cite{he2016deep,liu2022convnet,radford2021learning,dosovitskiy2020image,zhai2023sigmoid,zhai2022scaling,cherti2023reproducible,dehghani2023scaling,chen2024internvl,fini2024multimodal} and self-supervised pre-training~\cite{chen2020simple,he2020momentum,caron2021emerging,grill2020bootstrap,assran2023self,el2024scalable,chen2020generative,he2022masked,oquab2023dinov2} have successfully scaled up the size of vision model to billions of parameters and pre-training data to billions of images, which is proven effective in numerous downstream tasks~\cite{yang2024depth,radosavovic2023real,ramesh2022hierarchical,wu2023policy,juneja2023visual,juneja2024dino}. However, all these vision models are pre-trained at a low resolution (\eg, 384$\times$384) and unable to scale up to higher pre-training resolution due to the costly processing of full images. Previous work has studied the importance of scaling up input resolution in vision pre-training~\cite{tan2019efficientnet,tan2021efficientnetv2,dollar2021fast,bello2021revisiting} and attempted at incorporating high-resolution pre-training data~\cite{oquab2023dinov2,fang2023eva,liu2022convnet,liu2021swin}. However, they are mostly limited to tasks such as ImageNet classification~\cite{russakovsky2015imagenet} that are less sensitive to input resolution by nature, and none of these work has tried scaling pre-training to ultra-high resolution (e.g., over 1K resolution) which proves to be important for various real-world applications including MLLMs~\cite{shi2025we,internvl2,dong2024internlm,wang2024qwen2}. RADIO~\cite{Ranzinger_2024_CVPR,heinrich2024radio} uses a maximum pre-training resolution of 1024$\times$1024 although it does not explore CLIP-style pre-training and is pre-trained by distilling from other pre-trained vision encoders and thus is limited by the resolution of the teacher encoders. Another thread of work tries to improve fine-grained visual perception via localized contrastive learning~\cite{chen2024contrastive,bica2024improving,sun2024alpha} but none of these work scales up the pre-training resolution with the proposed approaches. In this work we scale up vision pre-training to 4K resolution and show the significant improvement on different MLLM benchmarks.

\minisection{High-resolution visual processing in MLLMs.} 
The success of LLMs~\cite{GPT4,touvron2023llama,OpenAI_ChatGPT,bai2023qwen} has spurred the progresses of MLLMs~\cite{liu2023llava,lin2024vila,shi2024eagle,wang2024qwen2,chen2024far,liu2024llavanext,li2024llava} with evolving recipe~\cite{liu2024nvila,fang2024vila,wei2021finetuned,tong2024cambrian} and functionality~\cite{guo2024regiongpt,huang2024lita,cheng2024spatialrgpt,Chen_2024_CVPR}. Recent work shows the importance of high-resolution perception in MLLMs~\cite{liu2024llavanext,chen2024far,shi2025we,li2024mini,xu2024llava} for tasks that require fine-grained detail perception. Different approaches to high-resolution image processing for MLLMs have been proposed, including splitting the high-res image into smaller tiles and processing each tile separately with a pre-trained low-resolution vision model~\cite{shi2025we,li2024monkey,liu2024llavanext,chen2024far}, using multiple vision models to encode an image at different resolution~\cite{li2024mini,shi2024eagle,luo2024feast,tong2024cambrian}, and directly using one single encoder to process high-res images~\cite{wang2024qwen2,vasu2024fastvlm}. However, most of these work does not scale up image resolution during the vision pre-training stage and directly processes high-res images with vision encoders pre-trained on low-res images, which is shown to be suboptimal~\cite{shi2025we}. Furthermore, all the previous methods exhaustively process every pixel in the high-res images without any saliency-based or prompt-aware region selection, leading to extreme inefficiency and rendering them unable to process ultra-high resolution such as 4K resolution. In contrast, \model scales up to 4K resolution during the pre-training stage while using a top-down patch selection mechanism that significantly improves the efficiency.

\minisection{Token pruning for vision encoders and MLLMs.} Redundancy in visual information often leads to inefficiency encoding in vision models and MLLMs. Previous work has explored improving the efficiency of vision encoders by pruning redundant vision tokens~\cite{rao2021dynamicvit,yin2022vit,chen2023sparsevit,verelst2020dynamic,bolya2022token}, although the models are all optimized for a specific task such as image classification and cannot adapt to general VQA tasks where token dropping should be based on the input prompt. AbSViT~\cite{shi2023top} learns to focus on different image regions based on any text prompt but it does not prune irrelevant tokens and is more computationally expensive. Other work has explored adaptively focusing on important regions for fine-grained image recognition~\cite{wang2020glance,wang2021not} although is still limited to low resolution. With the recent progress of MLLMs, visual token pruning methods for MLLMs have been proposed~\cite{chen2025image,zhang2024sparsevlm,yang2024visionzip,xing2024pyramiddrop} which removes redundant tokens based on either the self-attention in ViT or the text-image cross attention inside LLM. However, these methods still need to process the whole image in the ViT, resulting in large ViT latency especiall for high-res images and making it unable to scale up to 4K resolution. SEAL~\cite{wu2024v} processes high-res images in a search-and-focus style although the searching process is guided by LLM which adds on huge latency. \model, on the other hand, processes 4K resolution images with significantly improved efficiency and demonstrates superior performance over previous methods.

\section{Current Limitations and Future Directions}

\subsection{Current Limitations}

\minisection{Data.} Designs of the pre-training data are not extensively studied in this work due to limit of resources: \textbf{1) Quality of high-res images.} It is not enough to have images that are just high-res. They should contain rich details as well so the model has the incentive to learn fine-grained representations. This is supported by a preliminary experiment where we compare the effect of SA-1B and DataComp data in pre-training. Specifically, we separately remove the SA-1B and DataComp data from pre-training and find that removing SA-1B has larger impact on the performance despite its smaller size than DataComp. We hypothesize this is because the images in SA-1B usually contains dense or small objects while DataComp images usually have large or even single object in spite of the high resolution. In this work, we do not add any filtering on the richness of visual details in the images, which is necessary for future improvement.  \textbf{2) Quality of local bounding boxes of salient regions.} In this work, we detect salient regions that contain dense or small SAM masks. This heuristic can be improved, for example, by running OCR model to detect regions with texts for text-heavy downstream applications or by filtering out SAM masks that only contain textures but not meaningful objects. \textbf{3) Quality of local captions.} We use off-the-shelf MLLMs to generate local captions without any data cleaning or filtering. Future improvements can include borrowing the data filtering techniques from LLM synthetic data generation or using visual fact checkers~\cite{ge2024visual} to ensure accuracy of the generated captions. Furthermore, the quality of high-res SFT data for MLLM also needs improving. Currently we vanillaly generate high-res QA data from low-res data, which does not reflect the real distribution of high-res images in the wild. Manual or automatic curation of QA pairs on natural high-res images is better.

\minisection{Training.} In the current model, patch selection is supervised by labeled bounding boxes of salient regions. This can have several drawbacks. For example, since we first generate boxes and then generate captions from the boxes, it is possible the caption also corresponds to other regions outside the box but we treat those regions as negative when training patch selection. On the other hand, the current patch selection only select important regions and process all the scales in those regions, while it is usually the case that not all the scales need to be processed. Learning to select which scale to process is hard to be supervised because of lack of labeled data. All these issues can be addressed by using, for example, reinforcement learning instead of supervised learning where the model learns to find an optimal patch selection strategy itself in order to maximize the downstream performance.

\minisection{Model.} There are several defects in the current model design. First, when we run Stage 2 and Stage 3 of \model for multiple times, each group of high-res patches only have the context of the low-res patches but not high-res patches from other groups. This might affect \model's ability to model long-range correlation of high-res details. On the other hand, when selecting high-res patches, the selected patches can be scattered around the whole image and sparse within a local region. The sparsity of information affects the quality of local features. This is remedied by adding a soft locality constraint on patch selection, \eg, the model is encouraged to select more centralized and clustered patches. We verify this by adding gaussian smoothing on the selection score map such that patches around any high-score patch also have high scores and each group of selected patches is more clustered. We find this improves the performance when selecting all the patches, supporting our hypothesis. Nevertheless, this degrades the performance when selecting only parts of the patches, which is possibly because the selection score of salient objects is smoothed out if they are small.

\subsection{Future Directions}

\minisection{Beyond vision-language pre-training.} While \model follows the paradigm of CLIP-style vision-language pre-training, the same idea can be applied to any vision pre-training paradigm such as self-supervised pre-training. Taking DINOv2~\cite{oquab2023dinov2} as an example, the original pre-training objective is to improve the representation consistency of two different views of the image, and using the similar idea of localized processing, one can scale DINOv2 pre-training to 4K resolution by applying the DINOv2 objective to local regions. Furthermore, despite the recent progress on vision expert aggregation~\cite{shi2024eagle,tong2024eyes,Ranzinger_2024_CVPR,heinrich2024radio,li2024mini,liu2023prismer}, it is still an open question of how to combine vision-language and self-supervised pre-training by either joint or sequential training and how it affects performances of MLLM and other downstream tasks. Ideally, pre-training on 4K resolution images with both types of objectives can help the model learn both verbalizable and nonverbalizable features, for example, action-related features for robotic tasks.

\minisection{Beyond images.} \model processes 4K-resolution images by removing the redundant patches. Videos, especially high-res, high-fps, and long-form videos, contain much more redundancy and thus can benefit more from the similar idea of patch selection. Furthermore, learning such a patch selection mechanism can force the model to better grasp the world knowledge behind the videos such as physics and dynamics. Although video pre-training still faces lots of problems such as lack of video data and high-quality captions, such bottom-up or top-down spatiotemporal patch selection is a promising way to scale up video pre-training.

\section{Conclusion}

We propose \model that scales up CLIP-style vision pre-training to 4K resolution with a near-constant cost. \model learns high-resolution perception through local-global region-caption contrast. It encodes low-resolution global image and selectively processes only informative high-resolution regions. \model enables \vilamodel, a high-res MLLM whose performance scales with the \model pre-training resolution and surpasses state-of-the-art MLLMs with better efficiency. We also introduce 4KPro, a benchmark that evaluates visual perception at 4K resolution, on which \vilamodel outperforms state-of-the-art MLLMs.

\minisection{Acknowledgement.} We would like to thank Ekta Prashnani, Joohwan Kim, Alex Naumann, Subhashree Radhakrishnan, Vishwesh Nath, Yucheng Tang, Dong Yang, Holger Roth, and Daguang Xu for their assistance in data curation for the 4KPro benchmark. We are grateful to Greg Heinrich, Mike Ranzinger, Yi Dong, and Zhiding Yu for their feedback on the project. We thank Ligeng Zhu, Zhijian Liu, Yukang Chen, Yunhao Fang, and Xiuyu Li for helping with the codebase. We appreciate Long Lian, Junyi Zhang, and Haven Feng for discussion on the project. We thank Yin Cui and Ming-Yu Liu for their help with the benchmark data curation tools. We thank Andrew Tao, Kari Briski, Bryan Catanzaro, and Bill Dally for their feedback on the manuscript.

\newpage
{
    \small
    \bibliographystyle{ieeenat_fullname}
    \bibliography{main}
}

\clearpage
\appendix

\section{Details of \model Pre-Training Data Curation}
\label{appendix_sec:data_curation}

The full pre-training data sources and statistics are listed in Table~\ref{appendix_tab:data_statistics}.

\begin{table}[h]
\caption{\textbf{Data sources and statistics.} We collect in total 75M images with 1K - 4K resolution and 282M pairs of bounding boxes and detailed captions about salient local regions in the images.}
\label{appendix_tab:data_statistics}
\vspace{-0.5em}
\centering
\begin{scriptsize}
\begin{tabular}{lC{0.025\textwidth}cC{0.055\textwidth}C{0.025\textwidth}cC{0.055\textwidth}}
\toprule
\multirow{3}{*}{Data Source} & \multicolumn{3}{c}{1K - 2K Res.} & \multicolumn{3}{c}{2K - 4K Res.} \\
 & \#Img & \#Box & \makecell[c]{Avg. \\ Box size} & \#Img & \#Box & \makecell[c]{Avg. \\ Box size} \\
 \midrule
\multicolumn{3}{l}{\emph{Natural images}}\\
\ \ DataComp~\cite{gadre2024datacomp} & 18M & 54M & 424$\times$438 & 9M & 36M & 562$\times$578 \\
\ \ SA-1B~\cite{kirillov2023segment} & - & - & - & 11M & 44M & 302$\times$312 \\
\multicolumn{3}{l}{\emph{Documents}}\\
\ \ IDL~\cite{biten2022ocr} & 12M & 48M & 28$\times$286 & 7M & 28M & 30$\times$330 \\
\ \ PDFA~\cite{pdfa} & 12M & 48M & 80$\times$461 & 6M & 24M & 84$\times$569 \\
\midrule
\ \ Agg. & 42M & 150M & - & 33M & 132M & - \\
\bottomrule
\end{tabular}
\end{scriptsize}
\vspace{-0.5em}
\end{table}

\minisection{High-resolution images.}  The images consists of two types, natural images and documents. For natural images, we collect 18M images with 1K - 2K resolution and 20M images with 2K - 4K resolution from DataComp~\cite{gadre2024datacomp} and SA-1B~\cite{kirillov2023segment}. For documents, we take all 37M PDF pages from IDL~\cite{biten2022ocr} and PDFA~\cite{pdfa} and convert each page into image with DPI of 150, which normally results in images with resolution above 1.5K.

\minisection{Local captions and bounding boxes of salient regions for natural images.} In the saliency detection pipeline described in Section 2.1, we first use EfficientViT-SAM~\cite{zhang2024efficientvit} to generate all the masks in each image, similar to the ``segment everything'' mode in the original SAM~\cite{kirillov2023segment}. We use EfficientViT-SAM with model size of XL1. The arguments used for generating masks are listed in Table~\ref{appendix_tab:sam_args}. Notably, \texttt{points\_per\_side} and \texttt{crop\_n\_layers} largely affect how dense and detailed the generated masks are. 

After generating all the masks, we locate local bounding boxes of salient regions that contain small or dense masks. The detailed process is as follows: 1) We preset a set of boxes which are square, have the same sizes, and are uniformly distributed in the image with the distance between adjacent boxes equal to the size of the box. The box size is set depending on the size of the image. For example, for SAM, we set the box size as $1/5$ of the shortest side of the image. This results in a typical box size of 300 - 400 in a 2K resolution image. For each square box, we also preset two boxes at the same position with the same area as the square box but with aspect ratios of $1.5:1$ and $1:1.5$, respectively. 2) For each box, we calculate the saliency score of the box. The saliency score is the accumulation of the scores contributed by each mask that has overlaps with the box. The contribution to the score from a mask is calculated as $\frac{1}{\max(\texttt{Area}(mask), \ 40\cdot40)\ / \ \texttt{Area}(image)} \cdot \frac{\texttt{Area}(mask \ \cap \ box)}{\texttt{Area}(mask)}$, where the first term is larger when the area of the mask is smaller compared to the area of the whole image and the second term is larger when a larger portion of the mask resides inside the box. We then select the top-$k$ boxes with the highest saliency scores. To encourage the selected boxes to cover more areas, we ensure no overlap between boxes.

Finally, after detecting the local salient boxes, we generate captions about the local region by sending two images, the local crop and the global image, to an MLLM and asking it to generate a caption about the local details given the global context. Here we use Qwen2-VL~\cite{wang2024qwen2} as the MLLM since it has superior results to other open-source MLLMs and can handle multiple images. We set $\texttt{max\_pixels} = 256\cdot28\cdot28$ for Qwen2-VL to handle both the global image and the local crop. The prompt following the two images is: \textit{The second image is a crop from the first image. Given the context of the first image, please describe the second image briefly. Make sure to cover all the objects and texts in the second image. Make sure to describe all the attributes, including color, shape, spatial relations, of each object in the second image. If there's no text in the second image, you don't need to mention there's no text. Please only describe the objects in the foreground and don't describe the background such as the sky or the weather. Please only describe the objects in the second image and don't describe the objects in the first image. Please use 1-2 sentences.}

\begin{table}
    \caption{\textbf{Arguments for EfficientViT-SAM mask generation.}}
    \label{appendix_tab:sam_args}
    \vspace{-0.5em}
    \centering
    \begin{small}
    \begin{tabular}{lc}
        \toprule
        Argument & Value \\
        \midrule
        \texttt{points\_per\_side} & 24 \\
        \texttt{points\_per\_batch} & 128 \\
        \texttt{crop\_n\_layers} & 1 \\
        \texttt{crop\_n\_points\_downscale\_factor} & 1 \\
        \texttt{pred\_iou\_thresh} & 0.6 \\
        \texttt{stability\_score\_thresh} & 0.85 \\
        \texttt{min\_mask\_region\_area} & 0 \\
         \bottomrule
    \end{tabular}
    \end{small}
    \vspace{-0.5em}
\end{table}

\minisection{Local captions and bounding boxes for documents.} IDL provides bounding boxes and OCR results of each sentence, and PDFA provides the same labels for each word. For IDL, we randomly sample one sentence and use the bounding box as well as its OCR result as the caption. For PDFA, we sample 15 consecutive words and use the union of their bounding box and concatenate these words as the caption.

\minisection{Global captions.} We directly use Qwen2-VL to caption the whole image. The prompt is as follows: \textit{Please describe the image briefly. Make sure to cover all the objects and texts in the image. Make sure to describe all the attributes, including color, shape, spatial relations, of each object in the image. If there's no text in the image, you don't need to mention there's not text. Please only describe the objects in the foreground and don't describe the background such as the sky or the weather. Please use 2-3 sentences.}

\begin{figure*}[h]
    \centering
    \includegraphics[width=0.9\linewidth]{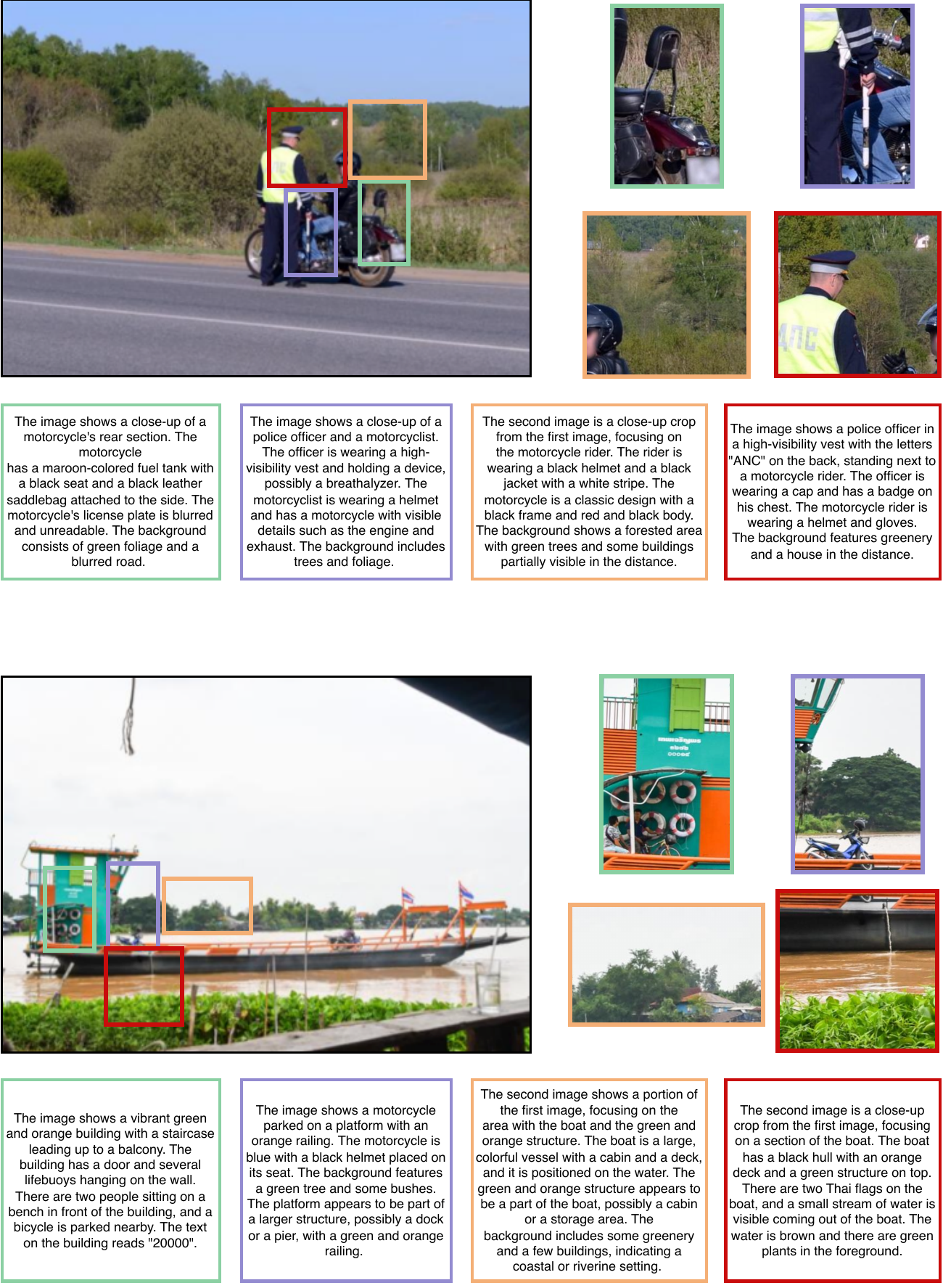}
    \caption{\textbf{Examples of pre-training data with natural images.} Here each image is labeled with bounding boxes of four salient regions (highlighted by different colors), together with the local captions of each region. The local captions, generated by Qwen2-VL, contains details in the crops although there are still occasional hallucinations.}
    \label{appendix_fig:pretrain_data_example_sam}
\end{figure*}

\begin{figure*}[h]
    \centering
    \includegraphics[width=0.9\linewidth]{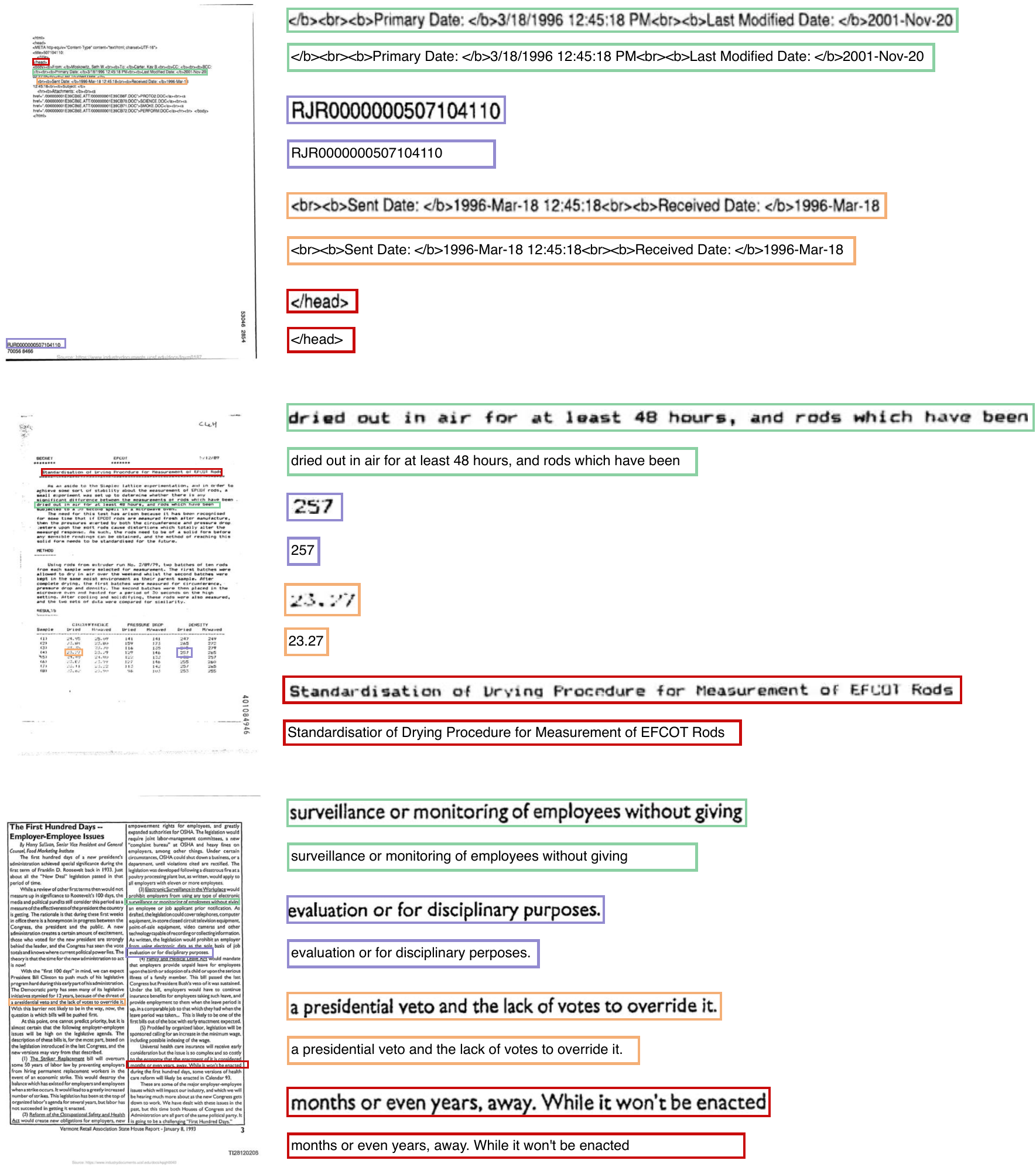}
    \caption{\textbf{Examples of pre-training data with document images.} Here each image is labeled with four bounding boxes (highlighted by different colors), together with the OCR results as the captions of each region.}
    \label{appendix_fig:pretrain_data_example_idl}
\end{figure*}

\minisection{Examples of the pre-training data.} See Figure \ref{appendix_fig:pretrain_data_example_sam}-\ref{appendix_fig:pretrain_data_example_idl}.

\section{Additional Details of \model Pre-Training Algorithm}
\label{appendix_sec:pretrain_detail}

During pre-training, since the data comes from different data sources (SAM, DataComp, IDL, and PDFA), we sample from each data source such that the probability each data source is sampled is the same for all data sources. Table~\ref{appendix_tab:hyperparameter_pretraining} shows the pre-training hyperparameters.

\begin{table}[t]
    \caption{\textbf{Hyperparameters of \model pre-training.}}
    \label{appendix_tab:hyperparameter_pretraining}
    \vspace{-0.5em}
    \centering
    \begin{small}
    \begin{tabular}{lc}
        \toprule
        Argument & Value \\
        \midrule
        \#epochs & 75 \\
        \#samples each epoch & 1e6 \\
        global batch size & 8192 / 4096 (for 3780 resolution) \\
        \%global caption in each batch & 25\% \\
        learning rate & 5e-6 \\
        warmup iterations & 1500 \\
        beta1 & 0.9 \\
        beta2 & 0.95 \\
        weight decay & 3e-4 \\
         \bottomrule
    \end{tabular}
    \end{small}
    \vspace{-0.5em}
\end{table}

\section{Additional Details of Training and Evaluation for \vilamodel}
\label{appendix_sec:mllm_training_setting}

\minisection{Detailed training and evaluation setting of Section~\ref{sec:scaling_property}.} The overall MLLM design and training pipeline of \vilamodel in Section~\ref{sec:scaling_property} follows NVILA~\cite{liu2024nvila}. Specifically, \vilamodel uses \model as the vision encoder, a two-layer MLP with 2$\times$2 spatial-to-channel reshaping as the projector, and Qwen2~\cite{yang2024qwen2} as the LLM. While the training pipeline of NVILA consists of five stages, we only use stage 1, 3, and 4 in training \vilamodel in this section, where stage 1 only trains the projector, stage 3 trains the trains the LLM and the projector, and stage 4 trains the whole model. For the training data, except for the self-collected data for patch selection and high-res feature alignment, we use the same stage 1 data as in NVILA, use only ShareGPT4V-Pretrain~\cite{chen2023sharegpt4v} for stage 3, and use a small subset of NVILA SFT data including ShareGPT4V-100K~\cite{chen2023sharegpt4v}, ShareGPT4V-SFT~\cite{chen2023sharegpt4v}, DVQA~\cite{kafle2018dvqa}, ChartQA~\cite{masry2022chartqa}, AI2D~\cite{kembhavi2016diagram}, DocVQA~\cite{mathew2021docvqa}, GeoQA~\cite{chen2021geoqa}, and SynthDoG-en~\cite{kim2022ocr} for stage 4 training. For training efficiency, instead of mixing the patch selection data with other data in the same stage of training, we use a separate stage to solely train on patch selection data. When training \vilamodel-4K, we adopt a dynamic-resolution training scheme, where we select high-res patches only at scales of 756 and 1512 and none of the patches at 3840 scale for images with resolution under 2646, and select patches for all three scales when the image resolution is above 2646, such that larger image regions are covered for low-res images while more 4K-resolution patches are selected for high-res images. The number of patches selected for each scale is proportional to the image area of that scale. For evaluation of \vilamodel-4K, we only select the high-res patches at 756 and 1512 scales for all benchmarks except for V$^\ast$ where we select patches from all three scales.

\minisection{Detailed training and evaluation setting of Section \ref{sec:4kpro} and \ref{sec:sota}.} The training setting in Section \ref{sec:4kpro} (the part of comparing to state-of-the-art MLLMs) and \ref{sec:sota} is roughly the same as Section~\ref{sec:scaling_property}, except that we adopt the first four stages in NVILA training pipeline instead of just three of them. The training data for each stage is the same as NVILA except that in stage 3 we only use 10\% of MMC4~\cite{zhu2023multimodal} data. For both \vilamodel-1.5K and \vilamodel-4K, we select 10240 high-res patches for each image during training, which is 70.2\% patches for \vilamodel-1.5K and 11.7\% patches for \vilamodel-4K. We limit the number of images for each instance to 4 during the first three stages and to 3 during stage 4 in order to improve training efficiency. During evaluation, for \vilamodel-1.5K selecting 44\% patches, the number of patches from 756 and 1512 scales are 2560 and 3840. For \vilamodel-4K selecting 18\% patches, we select all the patches from 756 and 1512 scales and none from 3840 scales for all the tasks except for high-res benchmarks which we select 1536, 6144, and 7680 patches for the 756, 1512, and 3780 scales. For \vilamodel-4K selecting 12\% patches, we select 2048 and 8192 patches for 756 and 1512 scales. For \vilamodel-4K selecting 6\% patches, we select 2048 and 3072 patches for 756 and 1512 scales. For \vilamodel-4K selecting 35\% patches on 4KPro, we select 1536, 6144, and 23040 patches for 756, 1512, and 3780 scales.

\section{Additional Comments on Training \vilamodel}

\textbf{Parallel training of next-token prediction.} Since the high-res patch selection is dependent on the latent embedding of the previous tokens, we cannot train next-token prediction for every token in parallel. Therefore, in training, we first run LLM on the low-res vision features and the text tokens to get the last-layer embedding of the last token, use it to extract high-res vision features, and then run LLM on the full sequence again for next-token prediction. Although this requires running LLM twice, we empirically observe the additional computational cost is marginal since the first run is on a shorter sequence and it does not require gradient because it is only for obtaining the last-layer embedding which is detached before the patch selection. Note that during inference the next token prediction is sequential so we only need to run LLM once for each token.

\section{Qualitative Examples of Patch Selection Fine-tuned with MLLMs}

\begin{figure*}[h]
    \centering
    \includegraphics[width=0.75\linewidth]{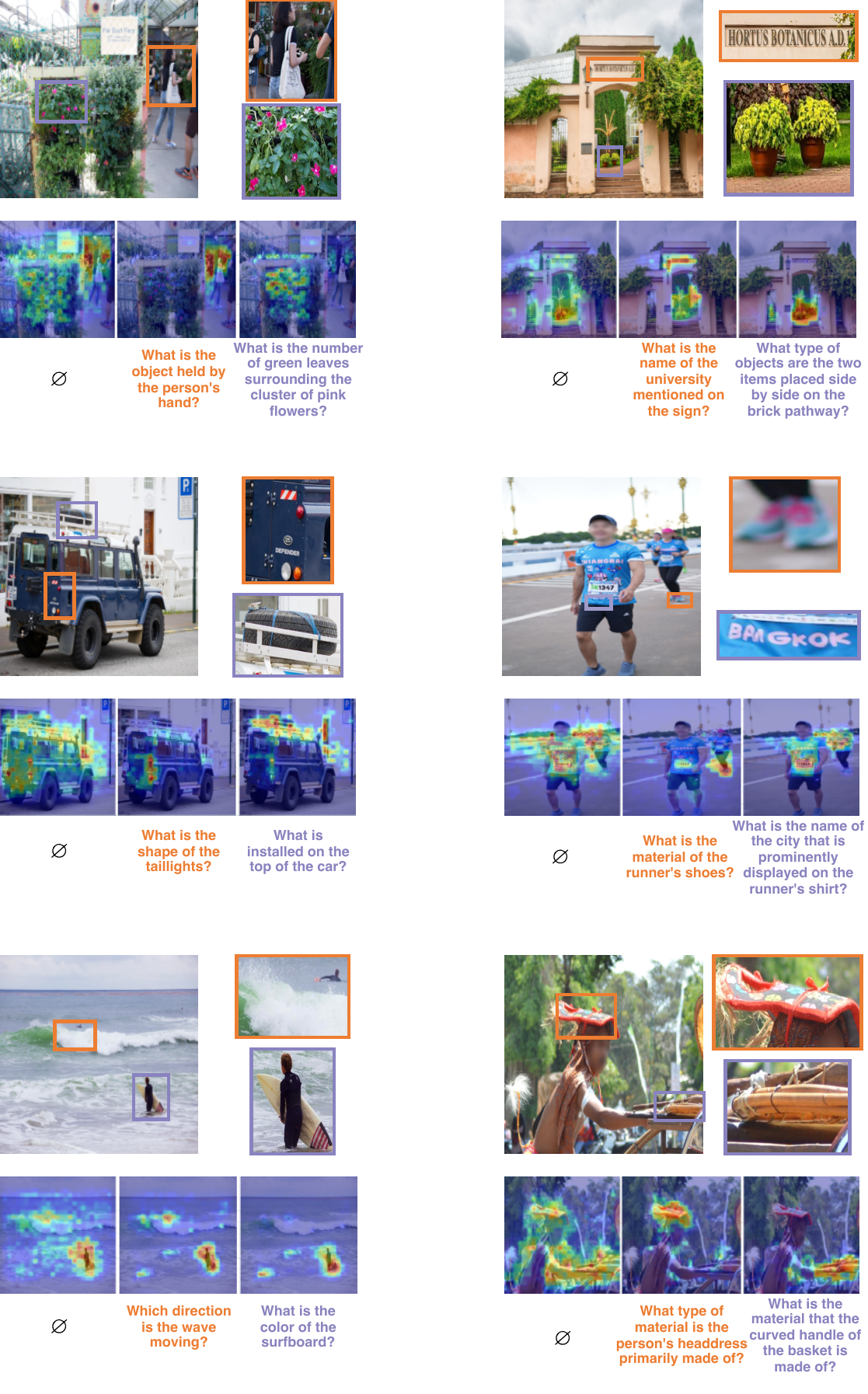}
    \caption{\textbf{Qualitative examples of patch selection on natural images.} \model is able to locate different parts of the image that are relevant to the question.}
    \label{appendix_fig:additional_qualitative_sam}
\end{figure*}

\begin{figure*}[h]
    \centering
    \includegraphics[width=0.6\linewidth]{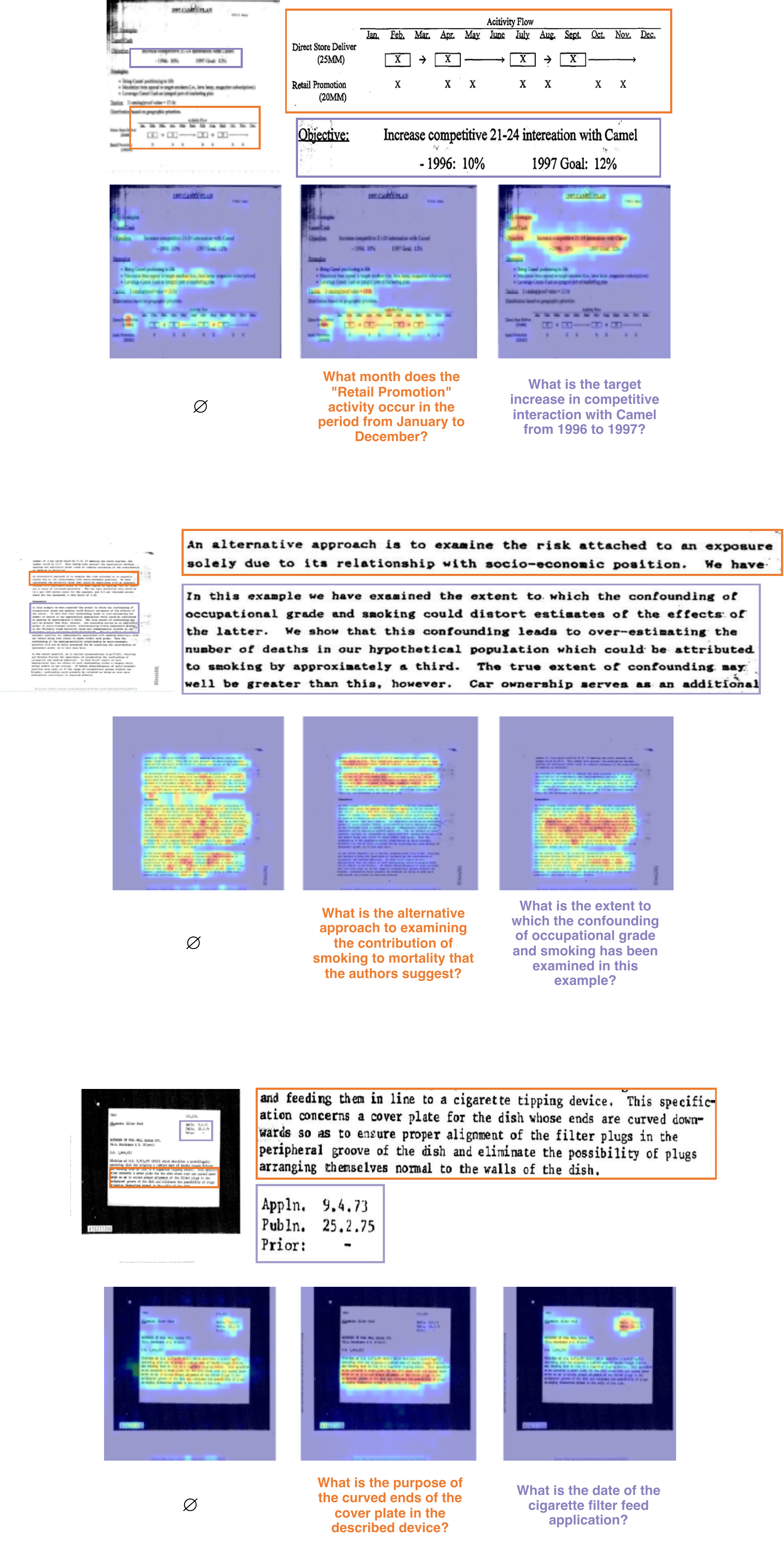}
    \caption{\textbf{Qualitative examples of patch selection on document images.} \model is able to locate different parts of the text in the document based on the question. For example, in the second example, when asked about the alternative approach of examination or the extent to which the confounding is examined, \model is able to locate the texts that discuss these topics.}
    \label{appendix_fig:additional_qualitative_idl}
\end{figure*}

We show additional qualitative examples of bottom-up and top-down patch selection of \model in Figure~\ref{appendix_fig:additional_qualitative_sam} and \ref{appendix_fig:additional_qualitative_idl}. 

\section{4KPro Data Curation}
\label{appendix_sec:4kpro_curation}

For each category, we first collect videos with 4K resolution (3840$\times$2160) from YouTube, and for each video, we manually select 5 frames containing rich details that are only recognizable under high resolution and label the bounding boxes around the local details. To obtain QA pairs about the details for each frame, we first use GPT-4o~\cite{hurst2024gpt} to generate a candidate QA pair based on the context of the global image and the content of the local crop, and then manually review and screen each generated QA pair to ensure each question is answerable under and only under 4K resolution and each answer is correct. %

\section{Additional Qualitative Results on 4KPro}

\begin{figure*}[h]
    \centering
    \includegraphics[width=1\linewidth]{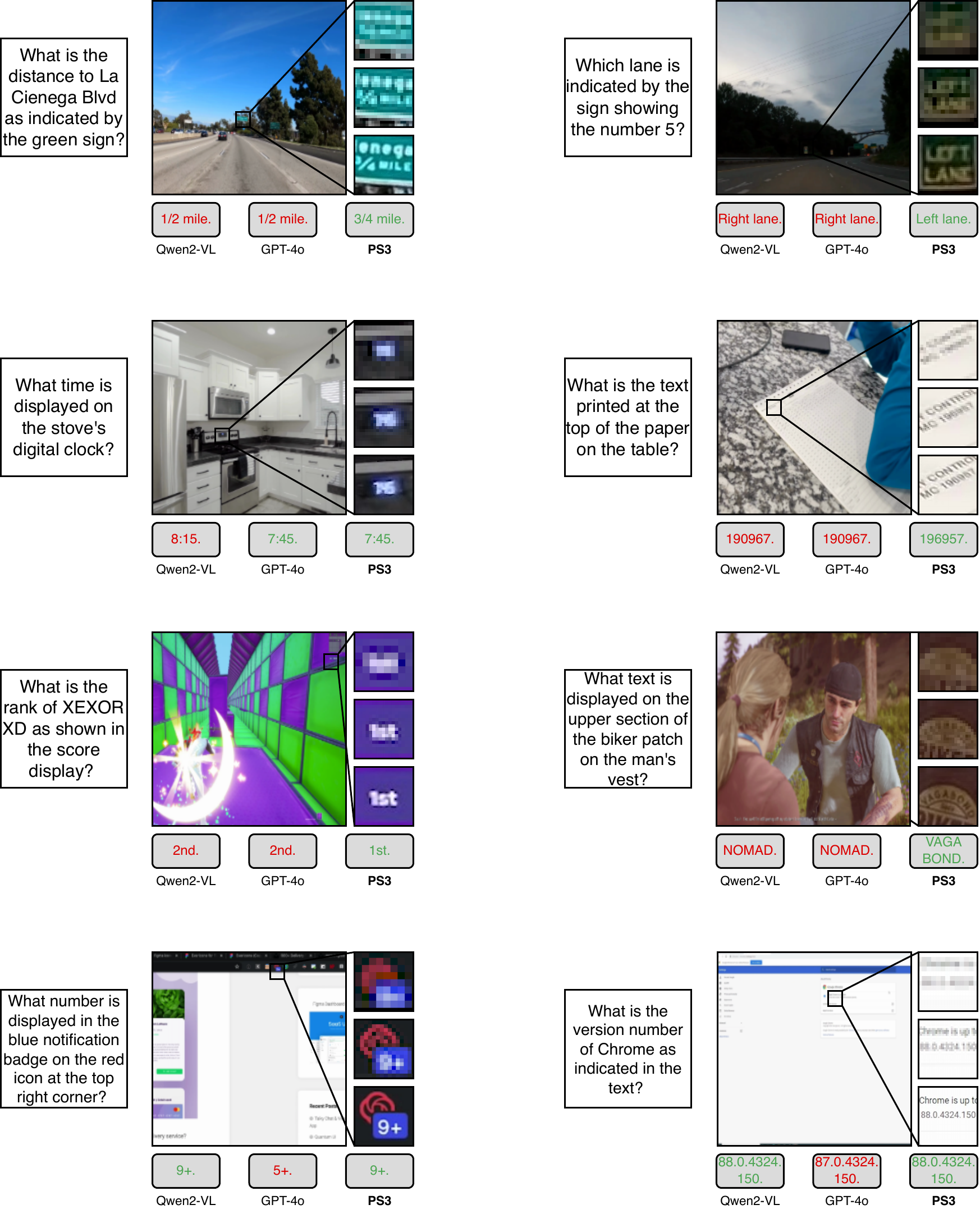}
    \caption{\textbf{Qualitative examples on 4KPro.} The four rows show examples from categories of autonomous vehicle, household, gaming, and UI understanding, respectively. For each instance, the local crop is shown under the resolutions of 756, 1512, and 3780.}
    \label{appendix_fig:additional_qualitative_4kpro}
\end{figure*}

We show additional qualitative comparison between \model and state-of-the-art MLLMs such as GPT-4o and Qwen2-VL in Figure~\ref{appendix_fig:additional_qualitative_4kpro}.

\section{Full Results of Scaling Properties on 4KPro}
Table~\ref{tab:scaling_4kpro} shows the full results of the experiment of scaling properties of \model on 4KPro (Figure 9 in Section 5.1).

\begin{table}[t]
\caption{\textbf{Full results of scaling properties of PS3 on 4KPro.}}
    \label{tab:scaling_4kpro}
    \centering
    \begin{small}
    \begin{tabular}{lllllc}
        \toprule
        Vision Encoder & \makecell[l]{Max \\ Res.} & \makecell[l]{\#HR \\ Token} & \makecell[l]{Select \\ (Train)} & \makecell[l]{Select \\ (Test)} & \makecell{Acc} \\ 
        \midrule
        SigLIP & 378 & 0 & - & - & 35.5 \\
        \midrule
        SigLIP-AnyRes & 756$^\dagger$ & 784$^\dagger$ & - & - & 37.1  \\
        SigLIP-\stwo & 756 & 784 & - & - & 40.7 \\
        PS3 & 756 & 729 & 100\% & 100\% & 41.9 \\
        \midrule
        SigLIP-AnyRes & 1512$^\dagger$ & 3136$^\dagger$ & - & - & 45.2 \\
        SigLIP-\stwo & 1512 & 3136 & - & - & 43.6 \\
        PS3 & 1512 & 3645 & 100\% & 100\% & 46.8 \\
        \midrule
        PS3 & 3780 & 1280 & 6\% & 6\% & 48.4 \\
        PS3 & 3780 & 3840 & 18\% & 18\% & 51.6 \\
        PS3 & 3780 & 7680 & 18\% & 35\% & 59.8 \\
        \bottomrule
    \end{tabular}
    \end{small}
\end{table}

\end{document}